\documentclass[twoside]{article}

\usepackage[accepted]{aistats2021}
\usepackage[round]{natbib}

\usepackage[utf8]{inputenc} 
\usepackage[T1]{fontenc}    
\usepackage{url}            
\usepackage{booktabs}       
\usepackage{amsfonts}       
\usepackage{nicefrac}       
\usepackage{microtype}      
\usepackage{multirow}
\usepackage{graphicx}
\usepackage{epstopdf}       

\usepackage{dblfloatfix}

\usepackage{microtype}
\usepackage{graphicx}
\usepackage{subfigure}
\usepackage{booktabs}
\usepackage[]{algorithm2e}
\usepackage{wrapfig}

\usepackage[usenames,dvipsnames]{xcolor}
\usepackage[colorlinks=true,citecolor=MidnightBlue]{hyperref}

\usepackage{amsbsy}
\usepackage{amsmath}
\usepackage{amssymb}
\usepackage{tikz}
\usetikzlibrary{fit,positioning}

\usepackage{pifont}

\newcommand{\cmark}{ \textcolor{green!60!black}{\ding{51}} }

\newcommand{\xmark}{ \textcolor{red!60!black}{\ding{55}} }

\def\f{\mathbf{f}}
\def\tf{\tilde{\mathbf{f}}}
\def\s{\mathbf{s}}
\def\a{\mathbf{a}}
\def\z{\mathbf{z}}
\def\0{\mathbf{0}}
\def\m{\mathbf{m}}

\def\v{\mathbf{v}}
\def\u{\mathbf{u}}
\def\K{\mathbf{K}}

\def\R{\mathbb{R}}
\def\E{\mathbb{E}}
\def\Z{\mathbf{Z}}

\def\N{\mathcal{N}}

\def\U{\mathbf{U}}
\def\bt{\boldsymbol{\theta}}
\def\S{\mathcal{S}}
\def\A{\mathcal{A}}
\def\cov{\mathbf{cov}}

\DeclareMathOperator*{\argmax}{arg\,max}

\begin{document}

%

%

\twocolumn[

\aistatstitle{Sample-Efficient Reinforcement Learning using Deep Gaussian Processes}

\aistatsauthor{ Charles Gadd\textsuperscript{1} \And Markus Heinonen\textsuperscript{1} \And Harri L\"{a}hdesm\"{a}ki\textsuperscript{1} \And Samuel Kaski\textsuperscript{1,2}}

\aistatsaddress{ \textsuperscript{1}Aalto University, Finland \And \textsuperscript{2}University of Manchester, UK}  ]

\begin{abstract}
Reinforcement learning provides a framework for learning to control which actions to take towards completing a task through trial-and-error. In many applications observing interactions is costly, necessitating sample-efficient learning. In model-based reinforcement learning efficiency is improved by learning to simulate the world dynamics. The challenge is that model inaccuracies rapidly accumulate over planned trajectories.  We introduce deep Gaussian processes where the depth of the compositions introduces model complexity while incorporating prior knowledge on the dynamics brings smoothness and structure. Our approach is able to sample a Bayesian posterior over trajectories. We demonstrate highly improved early sample-efficiency over competing methods. This is shown across a number of continuous control tasks, including the half-cheetah whose contact dynamics have previously posed an insurmountable problem for earlier sample-efficient Gaussian process based models.

\end{abstract}

\section{INTRODUCTION}

\begin{figure}
\centering
\includegraphics[width=0.35\textwidth]{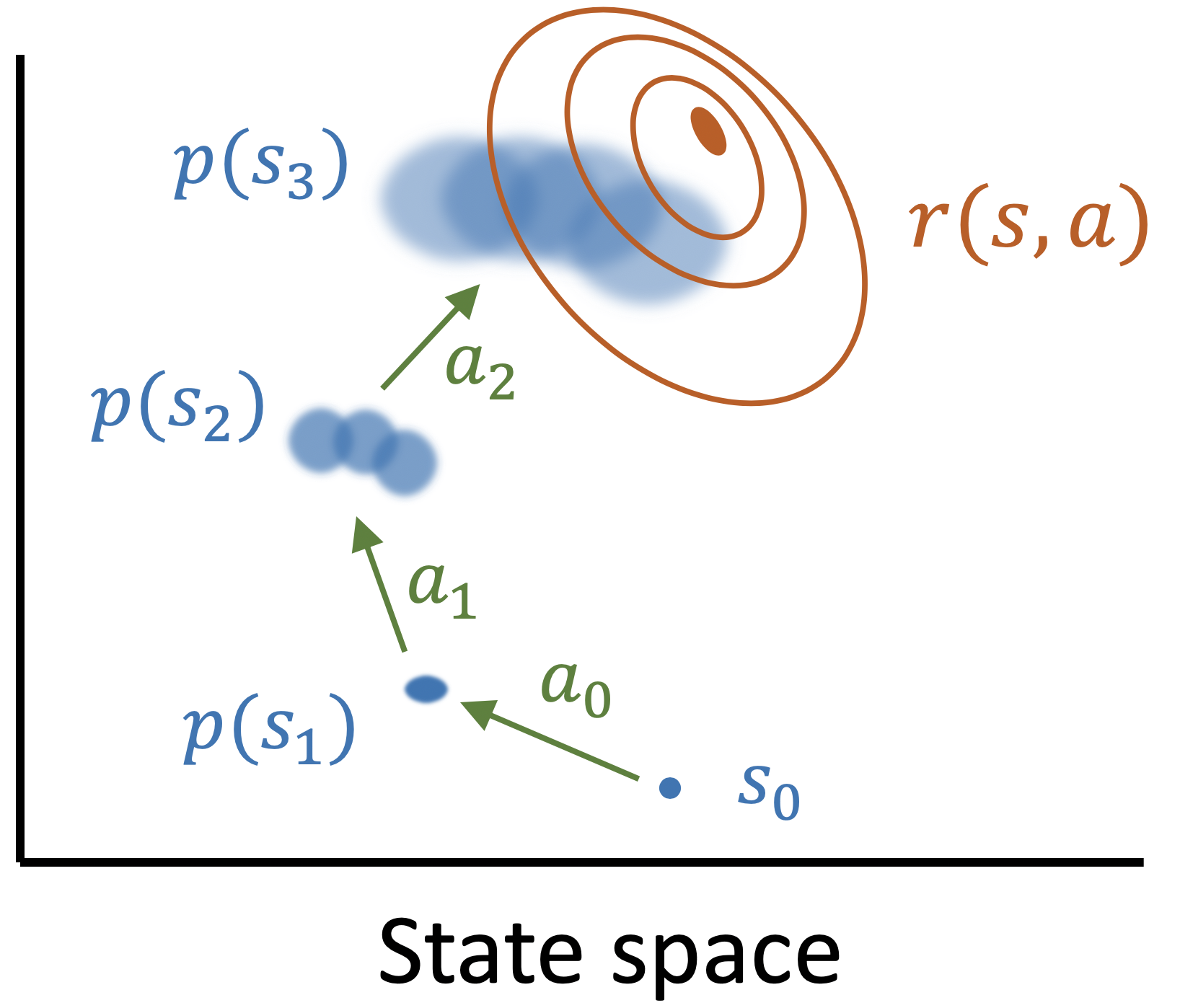}
\caption{Demonstration of how uncertainty in the states $\s_t$ progressively increases when simulating further into the future when designing actions $\a_t$ that maximise a reward $r$.}
\label{fig:cartoon_planning}
\end{figure}

Reinforcement learning (RL) provides a rigorous, automated framework for control and sequential decision making. Through trial-and-error, actions are chosen with the goal of achieving an objective defined \textit{a priori} which is often encompassed by a reward function. With each action taken our experience of interactions in a potentially complex environment grows, and our goal is to choose actions that maximise the reward. RL has been applied to numerous domains, ranging from robotics to gaming \citep{deisenroth2011pilco, kaiser2019model}.

Even in systems where both the environment and task are considered simple, many thousands or millions of observations are often required~\citep{sutton1998introduction}. In cases where simulation or experimentation is expensive, many RL approaches become impractical \citep{yu2018towards}. Recent advances in model-based reinforcement learning (MBRL) allow for sample-efficient RL by learning an environment dynamics model on much smaller data sets~\citep{chua2018deep, kamthe2017data, deisenroth2011pilco}. This model can then be used in proxy of a simulated or real experimental environment when learning an optimal policy or planning.

\begin{table*}[t]
    \caption{Model-based reinforcement learning methods.}
    \begin{center}
    \resizebox{0.8\textwidth}{!}{
    \begin{tabular}{l cccc r}
        \toprule
              & \multicolumn{2}{c}{Uncertainty} & & Exact & \\
        Model & Non-Gaussian & Non-stationary & Bayesian & sampling & Reference \\
        \midrule 
        PETS    &  \cmark  &  \cmark    & \xmark  & \xmark & \citep{chua2018deep} \\
        PILCO   & \xmark & \xmark & \cmark & \xmark & \citep{deisenroth2011pilco} \\
        DeepPILCO & \xmark & \cmark & \cmark & \xmark & \citep{gal2016improving} \\
        GP-MPC  & \cmark & \xmark & \cmark & \cmark & \citep{kamthe2017data} \\
        \midrule
        DGP-MPC & \cmark & \cmark & \cmark & \cmark & this work \\
        \bottomrule
    \end{tabular}
    }
    \end{center}
    \label{tab:methods}
\end{table*}

The choice of the dynamics model plays a crucial role. Its uncertainty or bias propagates through time, where uncertainty can explode within just a few trajectory steps (See Figure~\ref{fig:cartoon_planning}). Additionally, a sufficiently flexible model is required to model realistic non-smooth or non-stationary systems accurately. Crucially then, for sample efficiency we require a model with a principled framework for the inclusion of uncertainty, whilst maintaining the flexibility and capacity required to learn complex  systems.

Probabilistically principled models such as Gaussian processes are able to learn efficiently from very few observations, but they are limited in their ability to model complex systems due to their limited model capacity \citep{deisenroth2011pilco,kamthe2017data}. Conversely, the opposite is true for neural network (NN) based approaches which are able to model complex systems, but often require much larger datasets \citep{zhang2018study}. Uncertainty aware NN approaches have been developed with the aim of coupling the capacity of NNs with a probabilistic treatment of predictive uncertainty \citep{chua2018deep}. However, calibration of probabilistic NNs is still a challenging problem within MBRL \citep{guo2017calibration}.

We propose a model that combines the robust treatment of uncertainty of Gaussian processes, with the model complexity of deep learning. This leads to an approach that is able to rapidly learn even complex tasks with an extremely small number of samples. This is particularly useful in applications such as robotics, where contact points lead to complex dynamical systems, but observations of this system may be limited by factors such as experimental time, destructive testing, or economic cost~\citep{kamthe2017data}. We demonstrate significant improvements towards early learning efficiency in a number of OpenAI environments when compared with the current state-of-the-art.

\paragraph{Contributions.}
\textbf{1)} We introduce a novel deep, data-efficient model, \textbf{2)} which is both robust in the treatment of uncertainty, whilst simultaneously capable of modelling complex dynamics. \textbf{3)} We demonstrate our method exhibits strong early learning sample-efficiency over previous model-based and model-free methods on a number of benchmark environments. 


\paragraph{Related work.}
Gaussian processes are popular choices for model-based reinforcement learning due to their ability to effectively learn in low-data regimes whilst accurately capturing uncertainty when well-specified~\citep{kocijan2004gaussian, kuss2004gaussian, levine2011nonlinear, deisenroth2011pilco, kamthe2017data}. However, this has often come at the cost of strong prior assumptions which reduce the flexibility of the model~\citep{calandra2016manifold}. Consequently, applications have been largely limited to simple systems.

Both Bayesian and probabilistic neural network models have been proposed as the solution to the uncertainty modelling problem \citep{chua2018deep,gal2016improving}. While NNs are able to model complex dynamics, an inherent challenge in Bayesian Neural Network (BNN) modeling is to define meaningful priors especially for deep NNs with high-dimensional weight space. Additionally, approximate inference for BNNs is known to lead to poor uncertainty estimates~\citep{yao2019quality}. This behaviour is seen especially when making predictions away from the learning data distribution, where BNNs can give unsound and inaccurate predictions~\citep{sun2019functional}. Table~\ref{tab:methods} summarizes the key contributions implemented in our DGP-MPC approach, relative to previous methods.

\section{MODEL-BASED REINFORCEMENT LEARNING}

\begin{figure*}[t]
\centering
\begin{tikzpicture}[thick,scale=0.8, every node/.style={scale=0.8}]

    \tikzstyle{main}=[draw,circle,minimum size = 9mm,thick,text width=8mm,inner sep=0pt,align=center]

    \tikzstyle{connect}=[-latex, thick]

    \path
    (0, 1) node[main,fill=black!10] (s0) [] {$\s_0$}
    (1.5, 1) node[main] (s1) [] {$\s_1$}
    (4, 1) node[main] (sT1) [] {$\s_{H-1}$}
    (5.5, 1) node[main] (sT) [] {$\s_H$}

    (0, 2.2) node[] (a0) [] {$\a_0$}
    (1.5, 2.2) node[] (a2) [] {$\cdots$}
    (2.5, 2.2) node[] (a1) [] {$\a_{H-2}$}
    (4, 2.2) node[] (aT1) [] {$\a_{H-1}$}
    ;

    \path
    (s0) edge [connect] (s1) node[midway,below,xshift=20pt,yshift=30pt] {$\f$}
    (s1) edge [connect] (sT1) node[midway,below,xshift=75pt,yshift=30pt] {$\f$}
    (sT1) edge [connect] (sT) node[midway,below,xshift=160pt,yshift=30pt] {$\f$}

    (a0) edge [connect] (s1)
    (a1) edge [connect] (sT1)
    (aT1) edge [connect] (sT)
    ;

    \path
    (2.6, 1) node[circle,fill=white,minimum size=15pt,inner sep=0pt] (ell) [] {$\cdots$}
    ;

    \path
    (7, 2) node[] (at) [] {$\a_t$}
    (7, 0) node[main,fill=black!10] (st) [] {$\s_t$}
    (9, 1) node[main] (f1) [] {$\f_1$}
    (11, 1) node[main,text width=6mm,inner sep=0pt] (f2) [] {$\f_{L-1}$}
    (13, 0) node[main] (f3) [] {$\f_L$}
    (8.5, 3) node[main] (u1) [] {$\u_1$}
    (9.5, 3) node[] (t1) [] {$\bt_1,\z_1$}
    (10.5, 3) node[main] (u2) [] {$\u_{L-1}$}
    (11.9, 3) node[] (t2) [] {$\bt_{L-1},\z_{L-1}$}
    (12.5, 2) node[main] (u3) [] {$\u_L$}
    (13.75, 2) node[] (t3) [] {$\bt_L,\z_L$}

    (14.5, 1) node[] (b) [] {$\beta$}
    (16, 0) node[main,fill=black!10] (st1) [] {$\s_{t+1}$}

    (6.2, 0) node[] (hids) [] {}
    (16.8, 0) node[] (hid1) [] {}
    ;

    \path
    (st) edge [connect] (f1)
    (at) edge [connect] (f1)
    (b) edge [connect] (st1)
    (st) edge [connect,bend right=20] (st1)

    (f1) edge [connect] (f2)
    (f2) edge [connect] (f3)
    (f3) edge [connect] (st1)

    (u1) edge [connect] (f1)
    (u2) edge [connect] (f2)
    (u3) edge [connect] (f3)

    (t1) edge [connect] (f1)
    (t2) edge [connect] (f2)
    (t3) edge [connect] (f3)

    (hids) edge [connect,dashed] (st)
    (st1) edge [connect,dashed] (hid1)
    ;

    \path
    (9.9, 1) node[circle,fill=white,minimum size=15pt,inner sep=0pt] (ell) [] {$\cdots$}
    ;
    
    \draw[rounded corners=5pt,thick] 
    (7.9,-0.7) rectangle (15,4) node[xshift=-0.5cm,yshift=-8pt] {\textbf{DGP}}
    ;

\end{tikzpicture}
\caption{\textbf{Left:} The environment with fixed initial state $\s_0$ propagating into random variables $\s_1, \ldots, \s_H$ through dynamics $\f$, controlled by the optimized action parameters $\a_0, \ldots, \a_{H-1}$. \textbf{Right:} The Deep Gaussian process probability model. The first warping layers $\f_1, \ldots, \f_{L-1}$ are in the state-action space $\S \times \A$, while the last layer $\f_L$ maps to the state space $\S$. Inferred random variables are circled, while optimized hyperparameters are uncircled.
}
\label{fig:plate_diagrams}
\end{figure*}
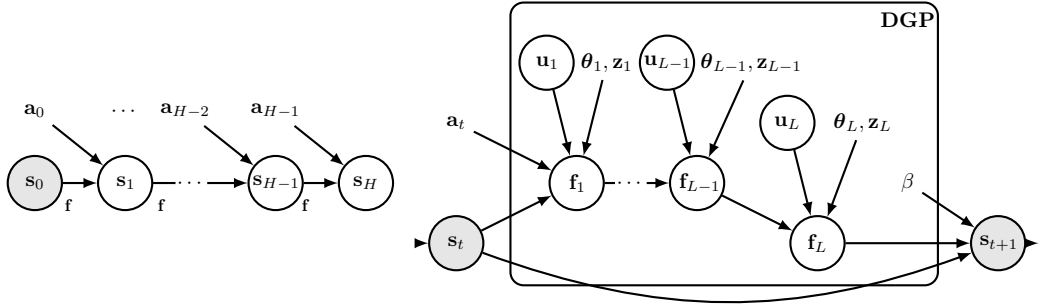

We consider interactions in an environment to be governed by a  discrete-time temporal dynamical system. Consider a deterministic dynamical system with states $\s_t\in\S \subset \R^{S}$ and actions $\a_t\in\A\subset\R^{A}$, indexed over discrete time points $t\subset\mathbb{N}$. We assume Markovian dynamics, with an underlying true state transition function that is unknown, deterministic and time invariant. We approximate this transition function with a stochastic function $\f : \S \times \A \mapsto \S$, which comes from a distribution of functions $p(\f)$. Here, the function $\f$ encodes the state transition 
\begin{align}
\mathbf{s}_{t+1} = \mathbf{s}_t + \mathbf{f}(\mathbf{s}_t,\mathbf{a}_t),
\end{align} 
as depicted in Figure~\ref{fig:plate_diagrams}. Additionally, we assume a control sequence $\a_{0:H-1} := (\a_0, \ldots, \a_{H-1})$ over the horizon $H$, which induces a joint distribution over state transitions
\begin{align}
    \label{eqn:jointtransition}
    p(\s_{1:H} | & \a_{0:H-1}, \s_0) = \prod_{t=0}^{H-1}  p(\s_{t+1} | \s_{t},\a_{t}) 
\end{align}
for a state sequence $\s_{1:H} := (\s_1, \ldots, \s_H)$, given an initial state $\s_0$. The state sequence distribution implicitly considers the distribution of transition dynamics $p(\f)$, resulting in stochastic trajectories. Given a fixed \textit{a priori} system-specific reward function, $r(\s_{t+1}, \a_t)$, the goal is to decide a control sequence $\a_{t:t+H-1}$ to perform at time $t$ in order to maximise the reward obtained.

We follow a sample-efficient model-based RL strategy of learning the transition dynamics $\mathbf{f}$ through emulation. Using a proxy dynamics model, transitions can then be sampled conditional on the current state and the action being considered. These transitions are then used to sample the expected reward for an action sequence under the model uncertainty. However, as we recursively sample transitions, the uncertainty of each pushes forward through the proxy model and results in diffuse state estimates. Consequently it is vital to then account for this uncertainty when performing optimal control, whilst maintaining the flexibility required to model non-stationary dynamics. This motivates a flexible and probabilistic emulator for the dynamical system.

\subsection{Modeling dynamics with deep Gaussian processes}

Given observed transition tuples $\left\{\s_t, \a_t, \s_{t+1}\right\}_t$ of the dynamical process, we approximate the transition function with a deep Gaussian process (DGP) \citep{damianou2013deep}, a composition of $L$ Gaussian process layers,

\begin{align} \label{eq:dgp}
    \s_{t+1} &= \s_t + \underbrace{\f_L}_{\S \times \A \mapsto \S} \circ \underbrace{\f_{L-1}  \cdots \circ \f_1(\s_t,\a_t)}_{\S \times \A \mapsto \S \times \A} + \epsilon,
\end{align}
where
\begin{align*}
    \f_\ell &\sim \mathcal{GP}(\m\left(\cdot\right), k_{\pmb{\theta}_\ell}\left(\cdot, \cdot\right) 
    ),
\end{align*}
with $\mathcal{GP}$ denoting a Gaussian process (GP) prior over function space, fully determined by a mean function $\m$ and kernel function $k$ with hyper-parameters $\pmb{\theta}_\ell$ \citep{williams2006gaussian}. A GP defines a distribution over functions $\f_\ell(\cdot)$ such that for any subset of inputs $\{ (\s,\a) \}$ the function outputs $\{ \f_\ell(\s,\a) \}$ follow a Gaussian distribution with a mean and covariance
\begin{align}
    \E[\f_\ell(\s,\a)] &= \m(\s,\a) \\
    \cov[\f_\ell(\s,\a), \f_\ell(\s',\a')] &= k_{\pmb{\theta}_\ell}((\s,\a), (\s',\a')).
\end{align}
In addition, $\epsilon$ is Gaussian additive noise with zero mean and a precision hyper-parameter $\beta$. This is included in our statistical model as a nugget on the last layer. 

In Equation \eqref{eq:dgp} the first $L-1$ layers $\f_\ell : \S \times \A \mapsto \S \times \A$ \emph{input-warp} the state-action pair $(\s_t,\a_t)$, and the final layer Gaussian process $\f_L : \S \times \A \mapsto \S$ maps into state transitions. The input-warping adds to our model its capability to estimate non-Gaussian uncertainty \citep{salimbeni2017doubly} over non-stationary transitions \citep{shen2019learning}.

\subsection{Sparse deep Gaussian process inference}

Following the sparse inducing point constructions of \citep{quinonero2005unifying,salimbeni2017doubly} for scalability, the Gaussian process transitions are guided by $M$ added inducing \emph{input} variables $\Z_\ell = \{ \z_{\ell,m} \}_{m=1}^M$ per layer consisting of state-action pair support vectors $\z \in \R^{S+A}$. Corresponding inducing \emph{outputs} $\U_{\ell} = \{ \f_\ell(\z_{\ell,m}) \}_{m=1}^M$ are also introduced that are in the state-action space for warping layers and in only the state space for final layer $L$. 

The joint distribution of the inducing DGP states and random variables at time $t$ follows
\begin{align} \label{eq:dgpjoint}
    p( \s_{t+1}, &\f, \U | \s_t, \a_t, \Z, \bt ) \\
    &= p(\s_{t+1} | \f_L, \beta) \prod_{\ell=0}^{L-1} p(\f_{\ell+1}|\f_\ell, \U_\ell, \Z_\ell, \bt_\ell) p(\U_\ell),  
\end{align}

with $\f_0 \triangleq (\s_t,\a_t)$,
\begin{align*}
     p(\f_{\ell+1}|\f_\ell, &\U_\ell, \Z_\ell, \bt_\ell) \\
    &= \N( \f_{\ell+1} | \K_{\f\z} \K_{\z\z}^{-1} \mathrm{vec} \U_\ell, \K_{\f\f} - \K_{\f\z} \K_{\z\z}^{-1} \K_{\z\f})
\end{align*}
and
\begin{equation*}
    p(\U_\ell) = \N(\mathrm{vec} \U_\ell | \m_{\z}, \K_{\z\z}),
\end{equation*}
where we use shorthands $\f = \{ \f_\ell \}$, $\U = \{ \U_\ell \}$, $\Z = \{\Z_\ell \}$, $\bt = \{ \bt_\ell \}$. 
The kernel matrices $\K_{\f\f}, \K_{\z\f}, \K_{\z\z}$ are evaluated at all pairs of corresponding sets $\f_\ell$ and $\Z_\ell$ with kernel parameters $\bt_\ell$. We use the ARD technique to estimate separate kernel bandwidths for each input dimension \citep{williams2006gaussian}. The first term of Equation~\eqref{eq:dgpjoint} is the Gaussian likelihood, while the second term is the kernel-based Gaussian process conditional of the next state. The final term is the inducing smoothness prior. Each output dimension is assumed to be independent, conditional on shared hyper-parameters. 

We follow the approach of \citep{havasi2018inference} to obtain posterior samples of $(\s_{1:H}, \f, \U)$ conditioned on observed trajectories with stochastic gradient Hamiltonian Monte Carlo (SG-HMC) inference, whilst estimating empirical Bayes values for $\bt$ and $\beta$ with a moving window approach. Further trajectories can be sampled efficiently from the posterior estimates of $(\f,\U)$ to obtain estimates of the trajectory uncertainty under the parameter uncertainty. We add jitter to all kernel inverses for numerical stability, and use separable, diagonal matrix-valued kernels \citep{micchelli2005learning}.

\section{PLANNING WITH MODEL PREDICTIVE CONTROL}

Having defined an approach for emulating the Markovian transition function that predicts the outcome $\f(\s,\a)$ of an action $\a_t$ at a state $\s_t$, this can now be used to perform control via planning to maximize the long term reward over a horizon $H$,
\begin{align}
    R(\a_{0:H-1}) = \sum_{t=0}^{H-1} \int r(\s_{t+1}, \a_{t}) p( \s_{t+1} | \s_{t}, \a_{t}) d \s_{t+1}.
\end{align}
Here, $r$ is a predefined reward of a state-action pair, and $p( \s_{t+1} | \s_t, \a_t)$ of Equations \eqref{eqn:jointtransition} and \eqref{eq:dgpjoint} is the recursive state distributions from an observed state $\s_{0}$. 
 We follow the Model Predictive Control (MPC) framework \citep{kamthe2017data,chua2018deep, mayne2000constrained}. Our goal is to find an action sequence over a horizon $H$ from time point $t$ that maximises the long term reward,
\begin{align}
 \hat{\a}_{t:t+H-1} = \argmax_{\a_{t:t+H-1}} R(\a_{t:t+H-1}).
\end{align}

Under MPC we then perform the first action $\hat{\a}_t$ of the sequence, before repeating the process again at $\a_{t+1}$. The long-term reward given an action sequence is an expectation over simulated trajectory samples $p(\s_{1:H} | \a_{0:H-1})$, whose variance arises from the uncertainty in the dynamics model $\f$.

To find the actions pertaining to the highest reward we employ the widely used cross-entropy method (CEM)~\citep{de2005tutorial}, which is an iterative stochastic optimisation procedure. Action sequences are sampled from a proposal distribution, the expected long term reward of each sequence is then predicted via sampling a trajectory under the model and the action sequences with the highest expected reward are used to refine the proposal distribution.

\subsection{Trajectory sampling}

We require a scheme in which we use the model to sample the posterior trajectory given a sequence of proposed actions,  Equation. We desire an approach which accurately captures model uncertainty and is not bound by restrictive distributional assumptions.

\citep{deisenroth2011pilco} used moment matching approximation, which maintains Gaussianity of the state $\s_t$ over time, which prevents estimating multimodal or skewed distributions. In \citep{chua2018deep} transitions are sampled from a bootstrapped model resulting in ensemble reward estimates. We propose an asymptotically exact particle-based trajectory sampling (TS) approach from the Bayesian model. We sample $P$ particles recursively and independently through the DGP-MDP
\begin{align}
    \s_{t+1}^{(p)} \sim \s_t^{(p)} + \tf_L \circ \tf_{L-1} \circ \cdots \circ \tf_1( \s_t^{(p)}, \a_t),
\end{align}
where $p=1,\dots,P$ is the particle index and the $\tf_\ell$ indicate the predictive posterior of each layer of our model. The predicted new state, from $\s_t$ under action $\a_t$ is then sampled from our model as $\s_{t+1}^{(p)}$. This procedure is carried out along the planning horizon, where each particle consists of recursive samples, and particles are independent of each other given the initial known state. 

These full-uncertainty sample trajectories are then pushed to the deterministic reward function. The long-term reward is the sum of rewards along this horizon, and the expectation under the model is used with the cross-entropy method to optimise the action sequence, as outlined in Algorithm \eqref{alg:ver2}.

\paragraph{Joint predictive posterior samples.} When planning, predictions for each particle are made using the joint predictive posterior distribution across all proposed action sequences simultaneously. The upshot of this is that, within each independent particle, trajectories remain correlated by state and actions. Consequently this smooths our reward objective. If two sequences are similar, or lead to a similar point in the state space then we expect the trajectories to follow consistent dynamics with each other.

\begin{algorithm}[t]
\KwData{Initial data $\mathcal{D}_0 = \left\{\s_t, \a_t, \s_{t+1}\right\}_t$ obtained by performing random actions. Initial action proposal distribution defined by $\m$ and $\v$.
}
\For{each episode $i$}{
  Train model: optimise $\left\{\Z, \theta, \beta\right\}$ and sample $\U$\\
  \For{each $t = 1, \ldots,$ TaskHorizon}{
    \For{each control optimisation iter}{
    Sample $K=300$ action sequences $\a^{(k)} \sim \N(\m, \mathrm{diag} \, \v)$ \\
    For each sequence, sample $P=5$ trajectories under the model\\ \hspace{1mm} $\s^{(p,k)}|\a^{(k)}$  \\
    For each sequence, calculate the expected long term reward over horizon $H$ \\ \hspace{1mm} $\sum_{\tau = t}^{t+H} \frac{1}{P}\sum_{p=1}^P r(\s_{\tau+1}^{\left(p,k\right)},\a_\tau^{(k)})$ \\
    Update $\m$ and $\v$ using the mean and variance of the top performing action sequences \\
    }
    Perform first action of the sequence $\a_t$ \\
    Append observed tuple $\left(\s_t, \a_t, \s_{t+1}\right)$ to data set.\\ 
    Re-sample inducing outputs $\pmb{U}$ \\
  }
 }
 \caption{Our DGP-MPC  algorithm.}
  \label{alg:ver2}
  \end{algorithm}

\section{EXPERIMENTAL RESULTS}

We now consider the performance of our model on a number of control tasks. We begin by first outlining these tasks, before analysing the improvements that are gained by increasing model-capacity through depth, and incorporating prior information through kernel choice. Finally we compare our approach to competing methods.

\subsection{Benchmark environments}

We consider three environments upon which to test our model. The first example is a modified cartpole, in which we demonstrate the advantages of improved model capacity even in simple settings. We then consider two additional environments: the reacher, in which the goal is to control a robot arm to reach a target; and the half cheetah environment, in which the goal to teach a cheetah to run. To the best of our knowledge this latter environment has previously posed insurmountable challenges for Gaussian process based models due to the contact points leading to significantly non-stationary dynamics. We demonstrate that increased model capacity allows us to tackle this problem. Additionally, we report significant sample efficiency improvements in the early stages of learning.

\begin{figure}
    \centering
    \subfigure[Cartpole]{
        \includegraphics[width=0.26\columnwidth]{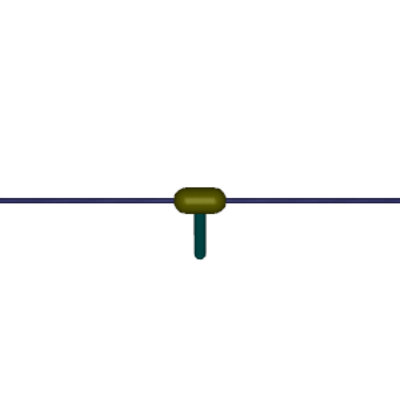}
        \label{fig:env_cartpole}
    }
    \subfigure[Reacher]{
        \includegraphics[width=0.26\columnwidth]{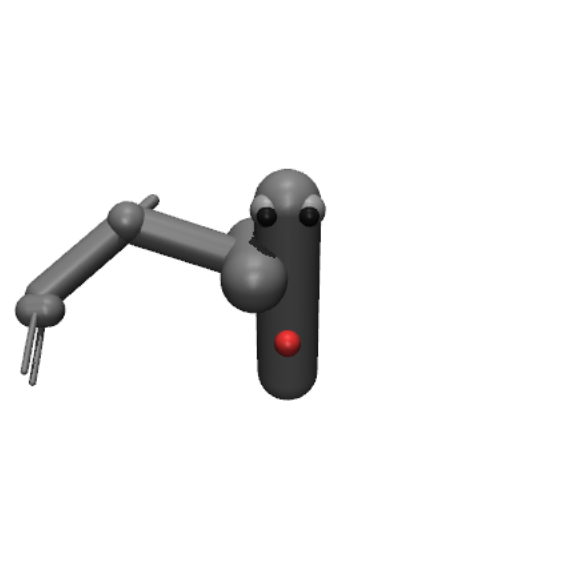}
        \label{fig:env_reacher}
    }
    \subfigure[Half-cheetah]{
        \includegraphics[width=0.26\columnwidth]{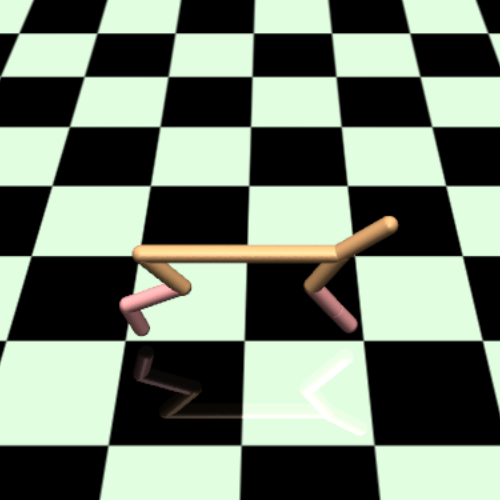}
        \label{fig:env_cheetah}
    }
\caption{Environments}
\end{figure}

\paragraph{Non-stationary cartpole.}
We consider the simple swinging cartpole task, as depicted in figure~\ref{fig:env_cartpole}, modified such that each episode begins with the cart at the end of the rail. Similar modifications have been considered in previous works with varying motivations,~\citep{kamthe2017data}. Within this task, the objective is to balance the pole above the cart, with a modification that the reward's target lies near the boundary.  Consequently, the true underlying dynamics are then non-stationary near the contact point at the end of the rail. In all cases reported in this example, where relevant we sample $K=300$ action sequences, use $200$ inducing points, and use $P=5$ particles. Each episode takes $200$ control iterations.

\paragraph{Reacher.}
We next consider the reacher environment, in which the objective is to direct a robot arm's end effector to reach a target ball, as depicted in figure~\ref{fig:env_reacher}. 
For this example we sample $K=300$ action sequences, use $200$ inducing points, and use $P=5$ particles.  Each episode takes $150$ control iterations

\paragraph{Half-cheetah.}
Lastly we consider the half-cheetah environment, depicted in figure~\ref{fig:env_cheetah}. Here our task is to make the cheetah run, where more reward is accumulated with greater speed. This environment poses a significant challenge for Gaussian process based models due to the increased number of dimensions and the highly non-stationary dynamics as a result of the contact points between the cheetah's feet and the floor. To the best of our knowledge no Gaussian process based dynamics model has yet been successfully applied to this environment in an RL setting. For this environment we sample $K=300$ action sequences, use $200$ inducing points, and use $P=5$ particles. We apply risk-sensitivity to control by penalising large back angles. Each episode takes $1000$ control iterations.

\begin{table*}[!b]
\centering
\begin{tabular}{ |p{3.5cm}||p{3.5cm}||p{2.5cm}|p{2.5cm}|p{2.5cm}|  }
 \hline
\multirow{2}{5em}{Environment}  & \multirow{2}{3em}{Kernel}  & \multicolumn{3}{|c|}{Average episode reward}   \\ 
& & $L=1$ & $L=2$ & $L=3$ \\
 \hline \hline
\multirow{4}{8em}{Modified Cartpole \hfill over $15$ episodes} &   Squared exponential   & $90.71 \pm 15.43$ & $132.73 \pm 7.10$ & $147.53 \pm 4.01$\\
 &   Mat\'{e}rn-$5/2$   &  $85.38 \pm 15.38$ & $133.36 \pm 6.61$  & $141.34 \pm 3.39$ \\ 
 &   Mat\'{e}rn-$3/2$ &  $95.97 \pm 15.59$ & $140.81 \pm 4.35$ & $149.81 \pm 3.76$ \\
  &   Mat\'{e}rn-$1/2$ &  $98.33 \pm 9.80$ & $141.98 \pm 4.86$ & $147.67 \pm 4.26$ \\
 \hline
 \multirow{4}{8em}{Reacher \qquad\qquad over $15$ episodes} &   Squared exponential  &  $-50.88 \pm 3.81$ & $-44.62 \pm 2.78$ & $-49.04 \pm 6.38$\\
 &   Mat\'{e}rn-$5/2$    & $-47.42 \pm 3.57$ & $-46.74 \pm 3.97$ & $-44.85 \pm 2.37$ \\
 &   Mat\'{e}rn-$3/2$    & $-45.09 \pm 3.53$ & $-44.39 \pm 1.90$ & $-44.68 \pm 3.14 $ \\
  &   Mat\'{e}rn-$1/2$   &  $-39.89 \pm 0.57 $ & $-40.82 \pm 1.26$ & $-41.28 \pm 2.39$ \\
 \hline
  \multirow{4}{8em}{Half cheetah \qquad over $10$ episodes} &   Squared exponential & $359 \pm 124.5$ & $1256 \pm 133.5$ &  $1498 \pm 70.4$\\
 &   Mat\'{e}rn-$5/2$    & $474 \pm 123.9$ & $1603 \pm 42.7$  & $1612 \pm 112.4$\\
 &   Mat\'{e}rn-$3/2$    & $671 \pm 83.9$ & $1540 \pm 139.5$ & $1680 \pm 83.3$\\
  &   Mat\'{e}rn-$1/2$   & $584 \pm 68.0$ & $1264 \pm 85.9$ & $1400 \pm 116.7$\\
 \hline
\end{tabular}
\label{tab:results}
\caption{Average episode reward for each environment under different kernel choices and model depth. Values reported are the mean and deviation over $10$ independent trials.}
\end{table*}

\subsection{Prior knowledge through functional priors.}

Through choice of kernel and depth, our Bayesian model leverages prior knowledge on the smoothness and non-stationarity of dynamical systems. In this section we study the effect of these model choices on performance.

\paragraph{Modelling contact dynamics.}

We first begin by motivating the use of deep Gaussian processes as an approach which accurately captures uncertainty, whilst incorporating prior information through kernel choice and increasing model capacity through depth. Here we consider the first modified cartpole environment. We train our DGP model on $50$ sequences of length $10$, with each sequence beginning at a random initial state. Sequences of positive actions are  sampled uniformly over $[0,3]$ and the resulting transitions are used to train our DGP model. Contact with the edge of the rail is then observed in all sequences. This is then tested by sampling the trajectory under a sequence of fixed actions with $a=1$ from the initial state mean, and comparing to the ground truth.

\begin{figure}
    \centering
    \subfigure[SEXP with $L=1$.]{
    \includegraphics[width=0.45\columnwidth]{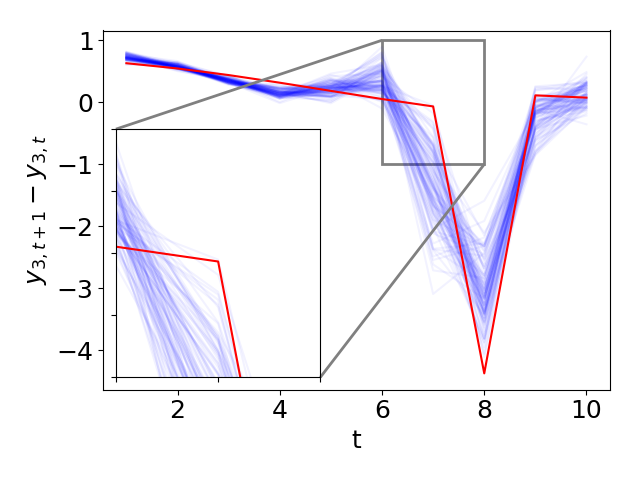}
}
\subfigure[SEXP with $L=2$.]{
    \includegraphics[width=0.45\columnwidth]{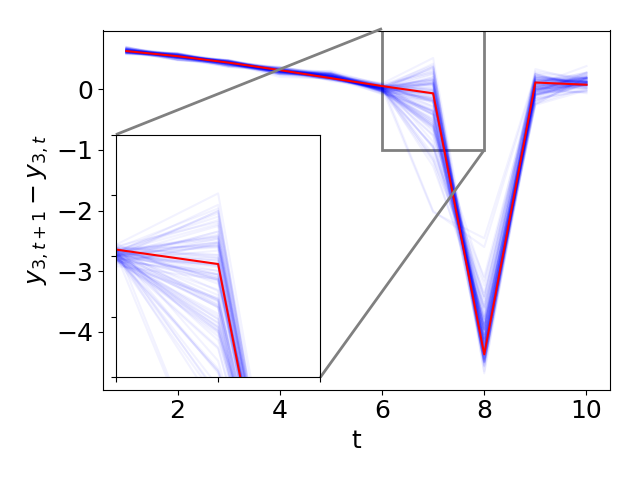}
}
\subfigure[Mat\'{e}rn-$\frac{3}{2}$ with $L=1$.]{
    \includegraphics[width=0.45\columnwidth]{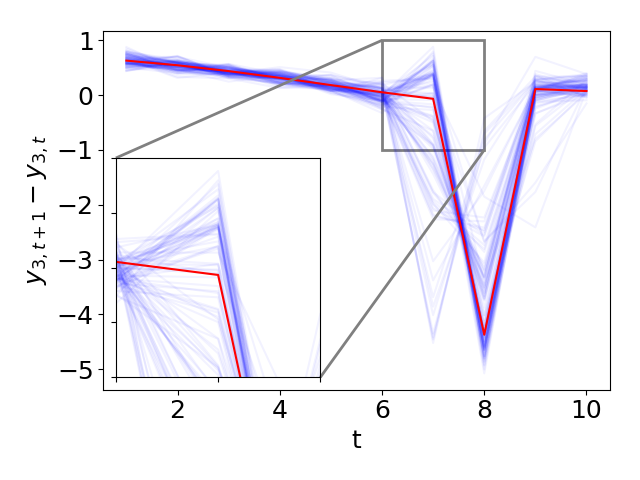}
}
\subfigure[Mat\'{e}rn-$\frac{3}{2}$ with $L=2$.]{
    \includegraphics[width=0.45\columnwidth]{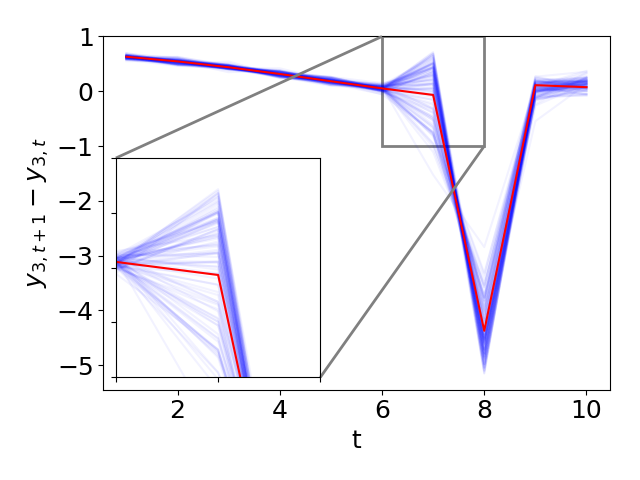}
}
\caption{Open loop trajectory samples for contact dynamics of the modified Cartpole environment. The change in state of the angular velocity is plotted against the time index of the sequence. Trajectory samples are shown in blue, whilst the true underlying dynamics are shown in red.}
    \label{fig:kernels_modified_cartpole_dynamics_crash}
\end{figure}

We consider the effect of increasing depth and choice of kernel in figure~\ref{fig:kernels_modified_cartpole_dynamics_crash}. Here posterior trajectory samples of Equation~\eqref{eqn:jointtransition} under our DGP model are shown in blue, whilst the true trajectory is shown in red. We clearly observe both improved predictive performance and more accurate uncertainty bounds  when using the once differentiable Mat\'{e}rn-$\frac{3}{2}$ kernel over the infinitely differentiable squared exponential (SEXP). Additionally, increased depth leads to further improvements as this allows the GP model to account for non-stationarity between regions of smooth and contact point dynamics. A more detailed study, including different scenarios, kernel choices, and numbers of layers is given in the supplementary material.

\paragraph{Prior knowledge with control.}

\begin{figure*}
     \centering
     \includegraphics[width=\textwidth, height=4cm]{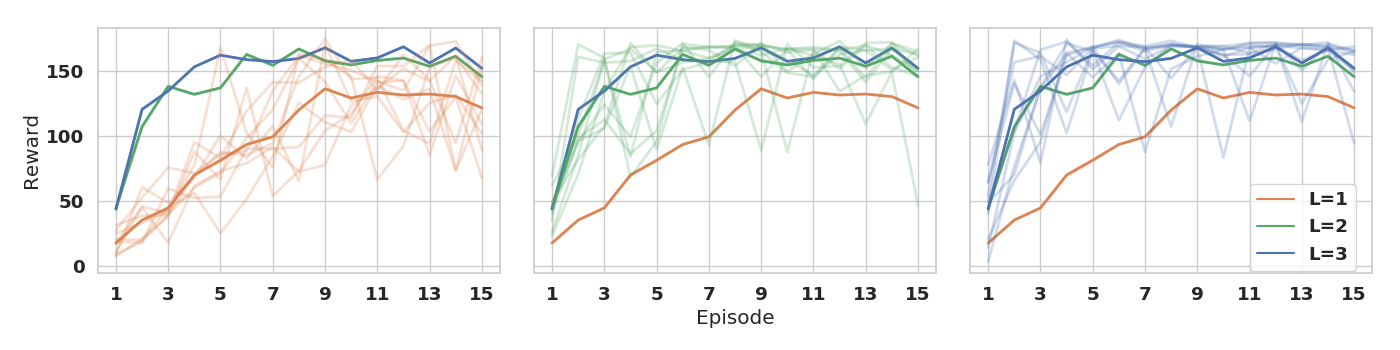}
     \caption{Modified cartpole reward under our approach with increasing number of layers $L$ and squared exponential kernels. Lines in the background  depict the reward observed across each independent experiment, whilst those in the foreground depict the mean observed reward across experiments.}
     \label{fig:cartpole_layers}
 \end{figure*}

We now study the effect of increasing the number of layers in the case of applying control. In figure~\ref{fig:cartpole_layers} we consider the modified cartpole environment, plotting the reward observed with increasing numbers of layers. Each experiment using our approach is repeated to account for stochastic effects. We observe a clear improvement as the number of layers increases. This is due to the increased ability to model the uncertainty of the non-stationary system whilst retaining accurate Bayesian uncertainty estimates. In addition, as the number of layers decreases, the model becomes increasingly miss-specified leading to larger predictive bias, increased variation across seeds, and consequently a less consistently performing controller. Similar results are seen across each environment.

Table~\ref{tab:results} reports the average obtained reward and deviation over independent trials. We choose this metric as it values both final performance and learning rate. Here, we consider the effect of incorporating prior knowledge on both depth and kernel smoothness. We observe that increasing depth greatly benefits those environments with non-stationary dynamics, whilst choosing an appropriately smooth kernel improves performance and improves stability. 

We note that whilst not used here, task-specific kernels have been shown to notably improve model specification even further~\citep{calandra2016manifold}. The ability to leverage this prior information within the kernels is a distinct advantage to our approach.

\subsection{Benchmark comparison}

We compare our approach to a number of competing methods. Firstly we consider \textit{PETS}, where the dynamics are modelled with ensembles of probabilistic neural networks and control is performed with planning via MPC. Two different trajectory sampling approaches are used: \textit{TS1}, in which a particle's bootstrap is continuously re-sampled; and \textit{TSinf}, in which a particle's boot strap does not change during a trial~\citep{chua2018deep}. To ensure a fair comparison throughout, all PETS experiments use comparable control parameters to our approach.  We also consider further Gaussian process approaches: \textit{GP-MM}, which approximately propagates the trajectory using moment matching; and \textit{GP-DS}, which approximately propagates the trajectory using distributional sampling. For these later GP experiments we again kept parameters comparable with our own choices.

\begin{figure*}[t]
     \centering
     \includegraphics[width=\textwidth, height=4cm]{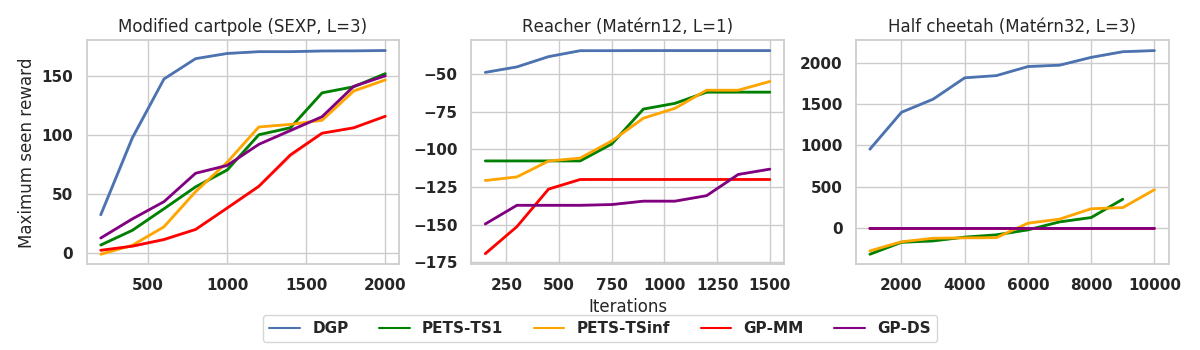}
     \caption{Maximum seen episodic reward, averaged over random seeds. Our approach using a DGP compared against a number of benchmarks.}
\label{fig:rewards}
 \end{figure*}

In figure~\ref{fig:rewards} the maximum seen rewards are plotted against the number of control steps. We report the result for our approach with $L=3$ and for each benchmark we report the mean over $10$ independent runs.  
When comparing our approach to each benchmark we clearly observe both a faster learning rate, and in many cases an improved final reward. In the half cheetah task we observe an exceedingly fast initial learning rate, with reward being accumulated from the very beginning given only a single episode of random interactions. In contrast GP-MM and GP-DS were unable to accumulate any reward, whilst PETS requires many more interactions before accumulating positive reward. As expected,  PETS does eventually surpass our performance, but requires many times more samples to do so. Plots showing performance with more episodes are given in the supplementary material, but lie beyond the scope of our sample-efficiency setting.

\begin{figure*}[t]
\centering
\subfigure[Cartpole]{
    \includegraphics[width=0.31\textwidth, height=4.25cm]{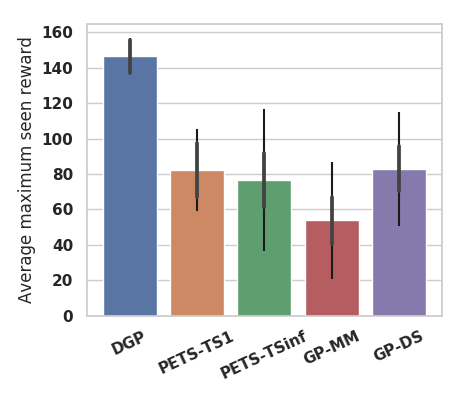}
    \label{fig:cartpole_variance}
}
\subfigure[Reacher]{
    \includegraphics[width=0.31\textwidth, height=4.25cm]{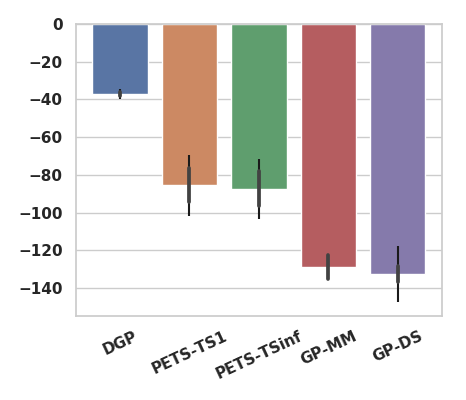}
\label{fig:reacher_variance}
}
\subfigure[Half-cheetah]{
    \includegraphics[width=0.31\textwidth, height=4.25cm]{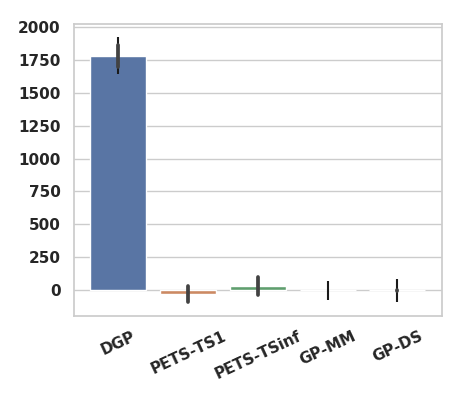}
    \label{fig:cheetah_variance}
}
\caption{Mean and deviation of the average reward obtained over the first $10$ episodes across independent trials. The larger the average reward, the quicker our agent converged to a strong optima.
}
\label{fig:variances}
\end{figure*}

We also consider the consistency of the learning process. Whilst the maximum seen reward gives a good depiction of the optimal performance of a controller through the exploration process, we are also concerned with the learning curves as depicted in figure~\ref{fig:cartpole_layers}, including their variability.
 For this we consider the variability across seeds of the mean reward across episodes.  The larger this mean reward, the faster a method led to a strong controller; the smaller the variability, the more consistently the model performs. This is shown in figure~\ref{fig:variances}. Variability of this estimator over seeds then also encapsulates erratic differences in rewards obtained between episodes, i.e. due to a controller which performs inconsistently. Our method's mean performance is clearly better than the alternatives in all cases, and variation is at most as large as the others, except for the half-cheetah where the others did not perform at all comparably.

\section{DISCUSSION}
Within this work we introduced a flexible, yet principled deep dynamics model-based  reinforcement learning algorithm for sample-efficient learning in non-stationary systems. We tested our algorithm on three control tasks, demonstrating significant learning efficiently and obtaining a strong performance within only a small handful of episodes. We demonstrated that incorporating prior knowledge and increasing model capacity through depth, whilst retaining robust uncertainty estimates, improved performance.

In future work a number of approaches can be implemented towards reducing the computational cost of calculating the predictive deep posterior \citep{wilson2020efficiently}. Towards this end, \citep{williams1998computation,khan2019} approximate the GP predictive posterior distribution with a neural networks or hybrid models while \citet{tsymbalov2019} approximates a neural network with a GP. These approaches aim to retain the benefits of probabilistic models while reducing the computational complexity. Similarly recent works of \citet{wang2019} and \citep{yu2019} improve Gaussian process numerics and inference, respectively.

Finally, our work on deep Gaussian process MBRL raises an interesting research question of combining the best of both worlds by learning ensembles of both neural and GP-based MBRL models with joint learning. In addition, recent work has focused on coupling model-based and model free approaches~\cite{nagabandi2018neural}. An interesting future avenue lies in coupling the sample efficiency of DGPs with the asymptotic properties of model-free approaches.

\newpage
\bibliographystyle{plainnat}
\bibliography{paper}

\end{document}


%

%

\onecolumn
\aistatstitle{Supplementary Material: Sample-Efficient Reinforcement Learning using Deep Gaussian Processes}

\section{Trajectory sampling - posterior predictive updates}
We propose an asymptotically exact particle-based trajectory sampling (TS) approach. Within this approach we place a composition of Gaussian process priors on a function space that models the transition from state-action pairs to the change in states. When performing trajectory sampling in the planning stage, transition samples are drawn from the predictive posterior of the Bayesian deep Gaussian process (DGP) model. Each trajectory sample is obtained by recursively and independently sampling through the posterior of the DGP-MDP model
\begin{align}
    \s_{t+1} \sim \s_t + \tf_L \circ \tf_{L-1} \circ \cdots \circ \tf_1( \s_t, \a_t),
\end{align}
where $\tf_\ell$ indicate the predictive posterior distribution of each layer of our model with a corresponding output sample $\tf_\ell^{(p)}$, where $p$ is the particle index. These samples are obtained through standard Gaussian process posterior prediction, leading to a joint distribution on latent functions across layers
\begin{align}
    (\tf_1^{(p)}, \dots, \tf_L^{(p)}) \sim p(\f|\u) = \prod_{\ell=1}^L p(\f_\ell | \f_{\ell-1, \u_\ell}),
\end{align}
where
\begin{align}
     p(\f_\ell | \f_{\ell-1, \u_\ell}) &= \mathcal{N}\left(\mu, \Sigma\right), \\
     \mu &= k\left(x^*, \Z\right)\left[k\left(\Z, \Z\right)+\beta_\ell^{-1}\I\right]^{-1} \U\\
     \Sigma &= k\left(x^*,x^*\right) - k\left(x^*, \Z\right)k\left[\left(\Z,\Z\right) + \beta_\ell^{-1}\I\right]^{-1}k\left(\Z, x^*\right)^T.
\end{align}
When considering the first layer, $x^*$ denotes the covariate inputs which are concatenated vectors of state and action, otherwise this denotes the latent function output of the previous layer. Each layer includes the homoscedastic precision term $\beta_\ell$, which is fixed to a small positive constant in all but the last layer.
The change in state $\s_t$ under action $\a_t$ is then approximated by our model. This procedure is carried out along the planning horizon, where each particle consists of recursive samples, and particles are independent of each other given the initial known state. Within each particle, predictive posterior samples are made jointly across all proposed action sequences.

\section{Cross Entropy Method}
The Cross Entropy Method (CEM) is a stochastic optimisation technique which refines a proposal distribution with the goal of maximising a noisy objective. In the context of our work, we have a proposal distribution  over action sequence and maximise the expected long term reward under our model.
 The CEM follows the procedure outlined in Algorithm~\ref{alg:CEM}. An in-depth tutorial can be found at~\cite{de2005tutorial}.

\begin{algorithm}
\KwData{Initial action proposal distribution defined by $\m$ and $\v$.
}
  \For{$i = 1,2,\ldots$}{
    Sample $K$ action sequences $\a^{(k)} \sim \N(\m, \mathrm{diag} \, \v)$ \\
    Obtain noisy observations of the reward via trajectory sampling \\
    Update $\m$ and $\v$ using the mean and variance of the top performing (elite) action sequences \\
    }
    \KwOut{An action sequence $\m$.}
 \caption{The CEM algorithm.}
  \label{alg:CEM}
  \end{algorithm}

\newpage
\section{Reproducing environments and experiments}
The parameters used throughout this work are reported in Table~\ref{tab:parameters_}. Where applicable, these choices are also used for the benchmark experiments.

\begin{table}[h] \centering
    \begin{tabular}{@{} cl*{7}c @{}}
        & & \rot{\shortstack[l]{Inducing\\points}} & \rot{Particles $(P)$} & \rot{Trajectories $(T)$}  & \rot{CEM elites} & \rot{CEM iterations} 
        & \rot{Horizon $(H)$} & \rot{\shortstack[l]{Actions performed\\per optimised sequence}} \\
        \cmidrule{2-9}
        & Modified Cartpole  &  200 &  5 &  300 & 10\% & 5 & 30  &  1 \\
        & Reacher     & 200 & 5 & 300 & 10\% & 5 & 20 &  1 \\
        & Half cheetah  & 200 &  5 & 300 &  10\% & 10 & 40  &  2\\
        \cmidrule[1pt]{2-9}
    \end{tabular}
\caption{Parameter choices for each environment}
\label{tab:parameters_}
\end{table}

\paragraph{Reacher. Target distribution.} Within the experiments reported the reacher target position distribution has been changed. The motivation is that under the original target distribution many sampled targets were not reachable under the physical constraints of the system, regardless of the controller performance. This change was implemented in the form of bias to the target distribution, with the new target distribution following a $\mathcal{N}\left([0.35, 0.05, 0], \text{diag}[0.05^2, 0.05^2, 0.1^2]\right)$ distribution. This modification was also applied to the benchmark environments for consistency. In addition to this, we seeded the target distribution samples to ensure fair comparison across runs and between methods.

\paragraph{Half cheetah. Risk-sensitive control.} For this environment we implemented risk-sensitive control by adding an additional term to the reward function \[ r\left(\s_{t+1}, \a_t\right) = s_{t+1, 0} - \left\lfloor\frac{|s_{t+1, 2}|}{\pi / 9}\right\rfloor - 0.1*\sum_i {a_{t,i}}^2,\] where $\s_{t+1, 0}$ is the velocity, $\s_{t+1, 2}$ is the back angle in radians, and $\a_t$ is the action performed, indexed by $i$. Whilst this term was included for control, reported rewards exclude this term for comparative purposes, instead reporting the original reward \[ r\left(\s_{t+1}, \a_t\right) = s_{t+1, 0}  - 0.1*\sum_i {a_{t,i}}^2. \]

\newpage
\section{Additional experiments}
Within this section we consider further complimentary experiments. Firstly, we consider the cartpole environment in the unmodified setting, in which the target of the end effector remains at the center point of the rail. Secondly, we consider the performance of the half-cheetah as the sample size increases beyond the scope of our sample-efficiency setting.

\subsection{Unmodified cartpole}

\begin{figure}[ht]
\centering
    \includegraphics[width=\textwidth]{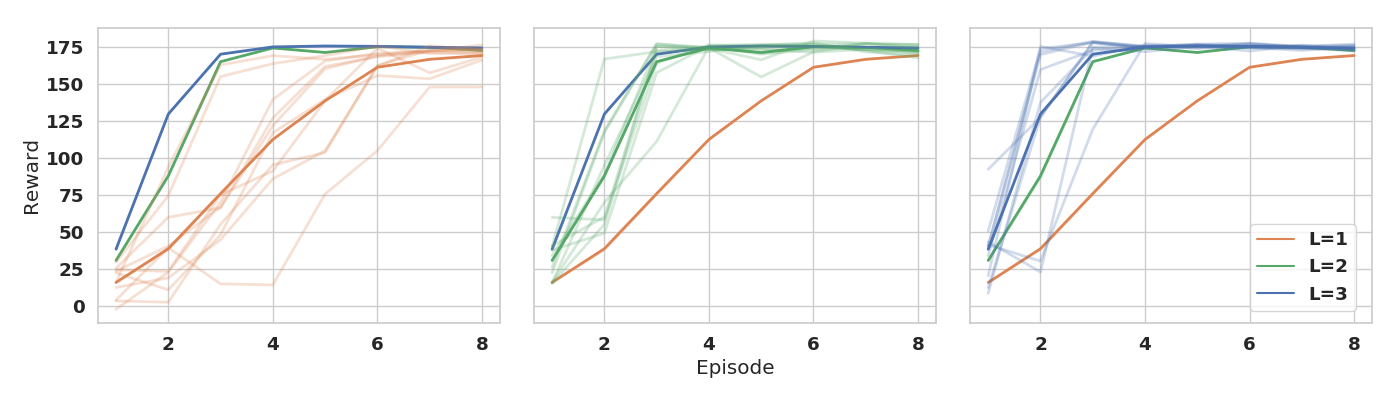}
\caption{The unmodified cartpole reward with increasing numbers of layers. Each case used the square exponential kernel across all layers.
}
\label{fig:unmodified_cartpole}
\end{figure}

We include here additional experiments for the unmodified cartpole example, in which the starting point remains at the center of the rail. Figure~\ref{fig:unmodified_cartpole} gives the maximum seen reward plots, in which the squared exponential kernel is used on every layer. We observe that we retain high sample efficiency in the simplified case.

\subsection{Half cheetah - beyond small sample sizes}

\begin{figure}[ht]
\centering
\includegraphics[width=\textwidth]{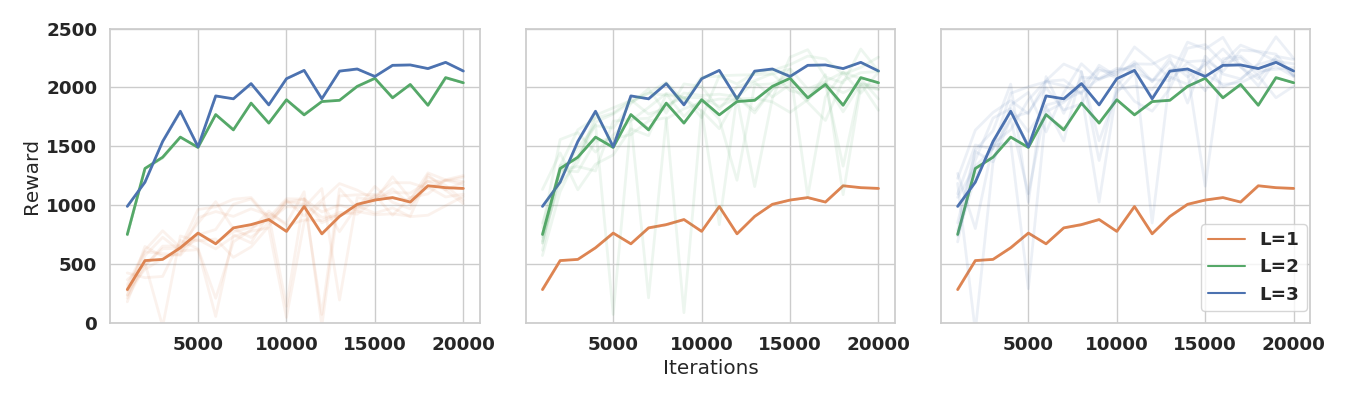}
\caption{The half cheetah reward with increasing numbers of layers. Each case used the Mat\'{e}rn-3/2 kernel across all layers..
}
\label{fig:cheetah_longer}
\end{figure}

We now consider the performance of our method when applied on the half cheetah, beyond the sample-efficiency domain. In Figure~\ref{fig:cheetah_longer} we present the reward obtained using our DGP-MPC approach for the half cheetah environment over a larger number of environment observations. 

\newpage
\subsection{Prior information. Dynamics of contact points.}
\def \kawFigPath {y0}

\subsubsection*{Modified cartpole} We study the effect of deep Gaussian process depth and kernel choice in the modified cartpole environment. Towards this end we consider three situations: a) an isolated non-stationary case of the cart hitting the wall b) an isolated stationary case of the cart travelling along the rail c) a hybrid case containing transitions of both. In each case the trajectory plotted is an open loop, where at each index the next sample is conditional on the previous sampled state, with no feedback. Here, steps are made on a finer time-scale to reduce discretisation errors. 

\begin{figure}[ht!]
\centering
\subfigure[SEXP kernel, $l=1$.]{
    \includegraphics[width=0.3\textwidth, height=3cm]{Images/supp/plots/dynamics/crash/SEXP_l1y3.png}
    \label{fig:}
}
\subfigure[SEXP kernel, $l=2$.]{
    \includegraphics[width=0.3\textwidth, height=3cm]{Images/supp/plots/dynamics/crash/SEXP_l2y3.png}
    \label{fig:}
}
\subfigure[SEXP kernel, $l=3$.]{
    \includegraphics[width=0.3\textwidth, height=3cm]{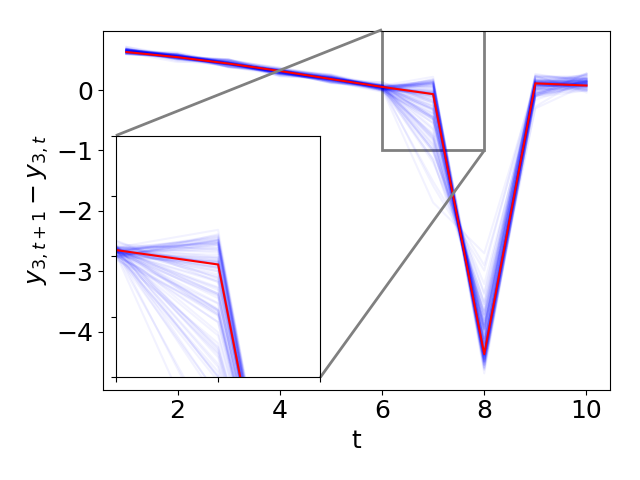}
    \label{fig:}
} \\
\subfigure[Matern52 kernel, $l=1$.]{
    \includegraphics[width=0.3\textwidth, height=3cm]{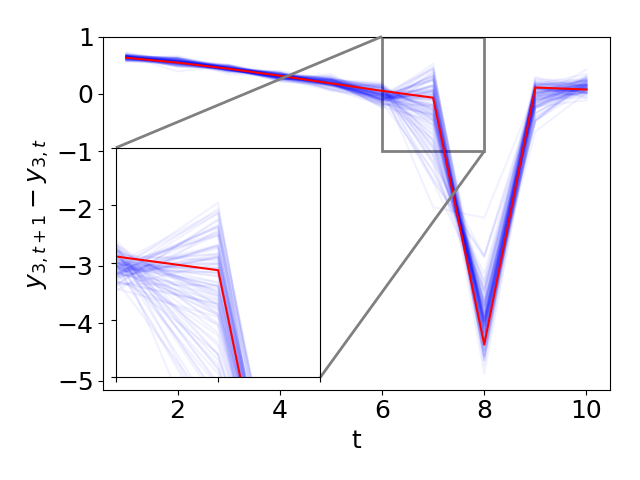}
    \label{fig:}
}
\subfigure[Matern52 kernel, $l=2$.]{
    \includegraphics[width=0.3\textwidth, height=3cm]{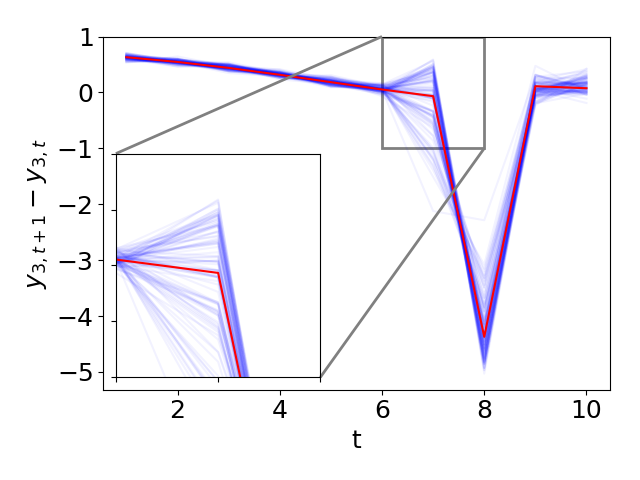}
    \label{fig:}
}
\subfigure[Matern52 kernel, $l=3$.]{
    \includegraphics[width=0.3\textwidth, height=3cm]{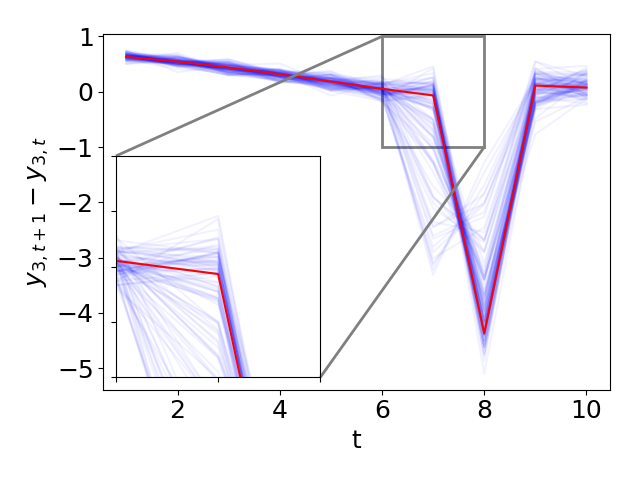}
    \label{fig:}
} \\
\subfigure[Matern32 kernel, $l=1$.]{
    \includegraphics[width=0.3\textwidth, height=3cm]{Images/supp/plots/dynamics/crash/Matern32_l1y3.png}
    \label{fig:}
}
\subfigure[Matern32 kernel, $l=2$.]{
    \includegraphics[width=0.3\textwidth, height=3cm]{Images/supp/plots/dynamics/crash/Matern32_l2y3.png}
    \label{fig:}
}
\subfigure[Matern32 kernel, $l=3$.]{
    \includegraphics[width=0.3\textwidth, height=3cm]{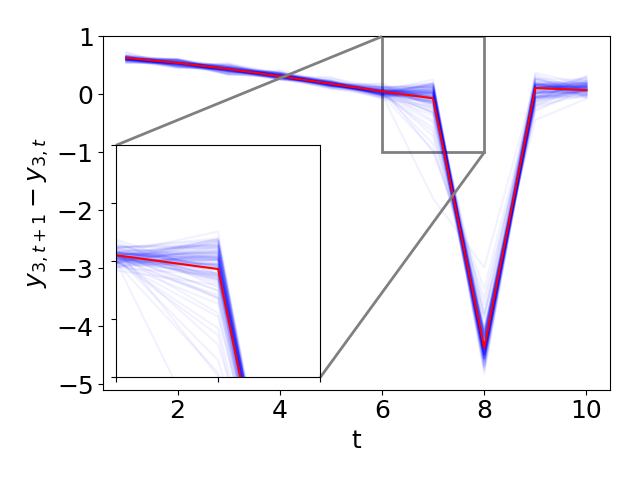}
    \label{fig:}
} \\
\subfigure[Matern12 kernel, $l=1$.]{
    \includegraphics[width=0.3\textwidth, height=3cm]{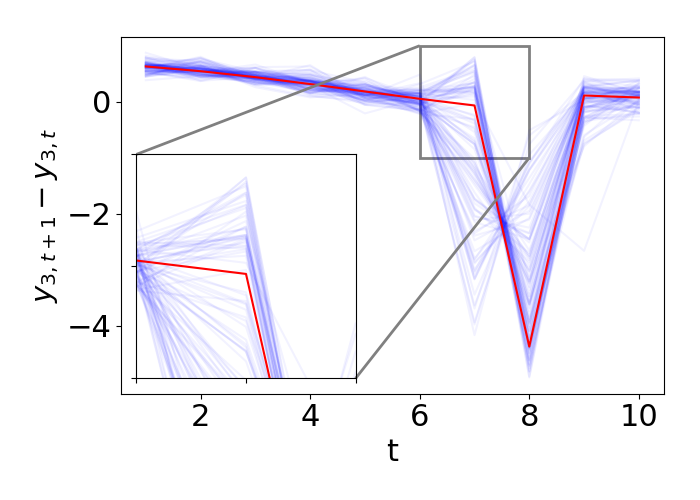}
    \label{fig:}
}
\subfigure[Matern12 kernel, $l=2$.]{
    \includegraphics[width=0.3\textwidth, height=3cm]{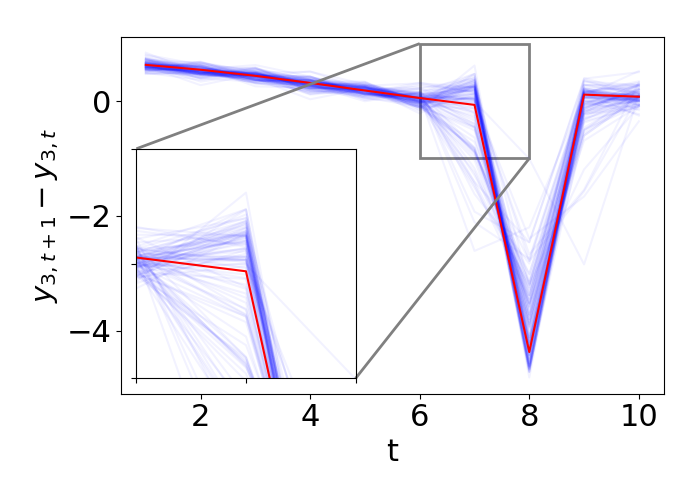}
    \label{fig:}
}
\subfigure[Matern12 kernel, $l=3$.]{
    \includegraphics[width=0.3\textwidth, height=3cm]{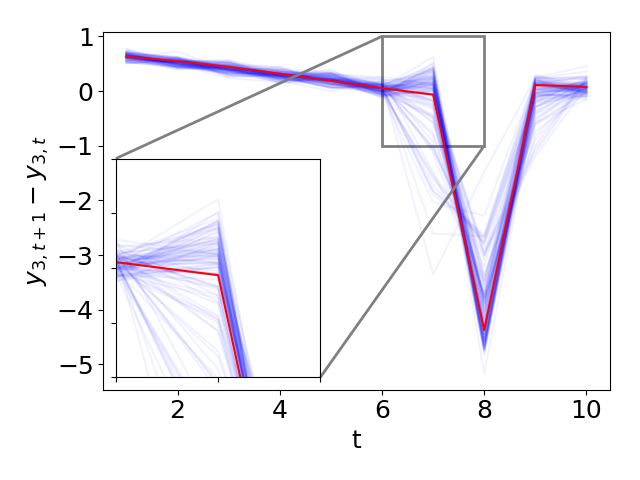}
    \label{fig:}
} \\
\caption{\textbf{Case a}. Analysis on the effect of kernel choice and depth when modelling the non-stationary contact dynamics in the modified cartpole example.
}
\label{fig:kernels_modified_cartpole_dynamics_crash}
\end{figure}

\paragraph{Case a} We train our DGP model on $50$ sequences of length $10$. Each sequence begins at a random initial point, under the same distribution used in our control experiments. Sequences of positive actions are then uniformly sampled in $[0,3]$ and the resulting transitions are used to train our DGP model. This is then tested by sampling the trajectory under a sequence of fixed actions with $a=0.75$ from the initial state mean, and comparing to the ground truth.  See Figure~\ref{fig:kernels_modified_cartpole_dynamics_crash}.

\begin{figure}[ht!]
\centering
\subfigure[SEXP kernel, $l=1$.]{
    \includegraphics[width=0.3\textwidth]{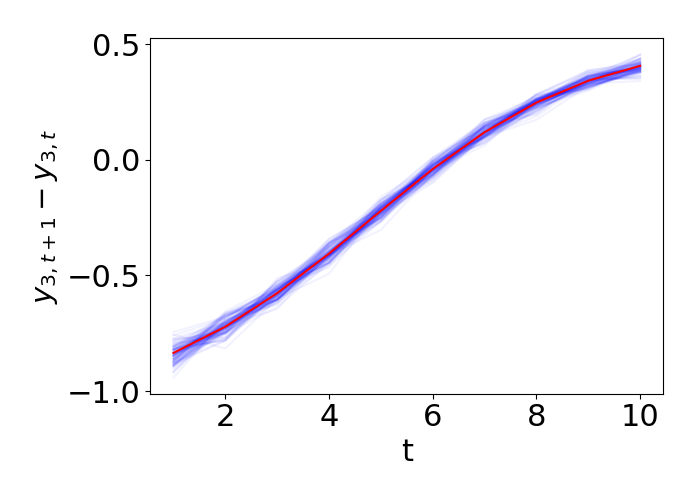}
    \label{fig:}
}
\subfigure[SEXP kernel, $l=2$.]{
    \includegraphics[width=0.3\textwidth]{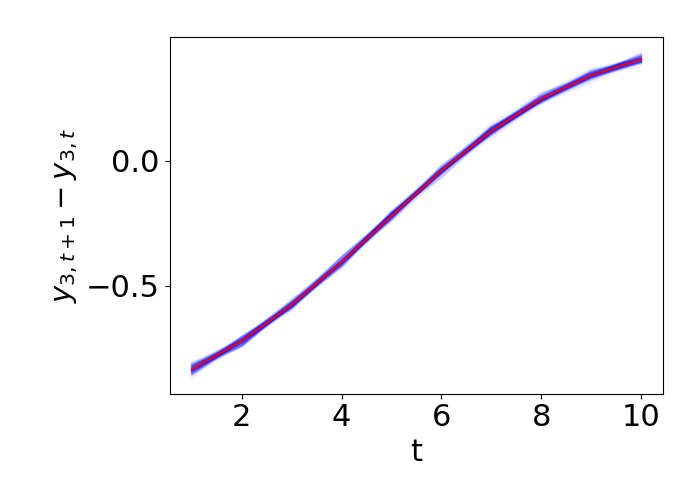}
    \label{fig:}
}
\subfigure[SEXP kernel, $l=3$.]{
    \includegraphics[width=0.3\textwidth]{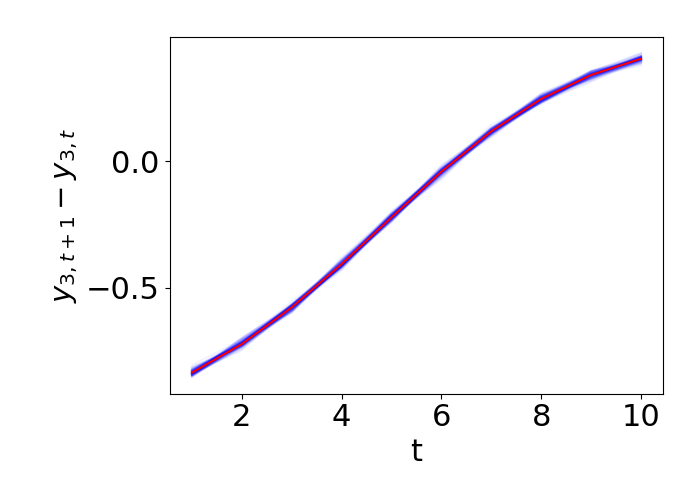}
    \label{fig:}
} \\
\subfigure[Matern52 kernel, $l=1$.]{
    \includegraphics[width=0.3\textwidth]{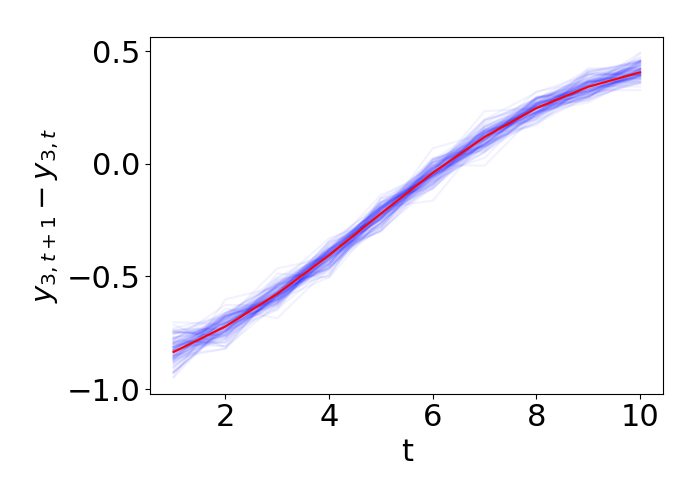}
    \label{fig:}
}
\subfigure[Matern52 kernel, $l=2$.]{
    \includegraphics[width=0.3\textwidth]{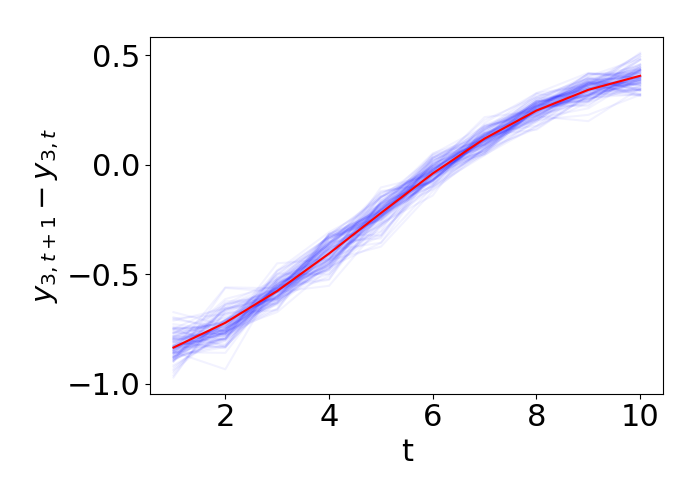}
    \label{fig:}
}
\subfigure[Matern52 kernel, $l=3$.]{
    \includegraphics[width=0.3\textwidth]{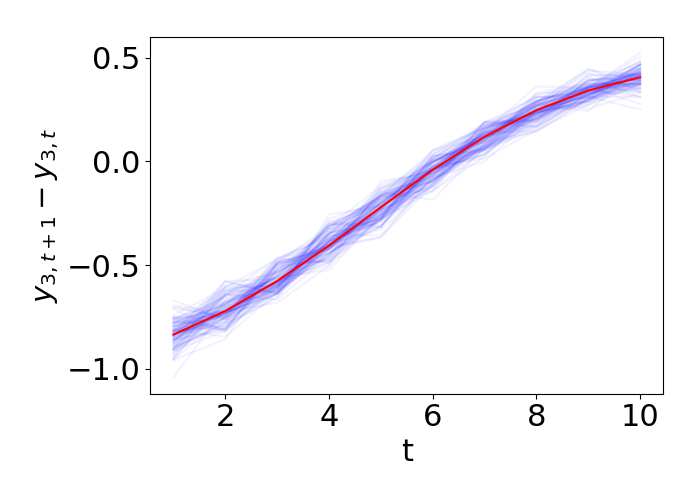}
    \label{fig:}
} \\
\subfigure[Matern32 kernel, $l=1$.]{
    \includegraphics[width=0.3\textwidth]{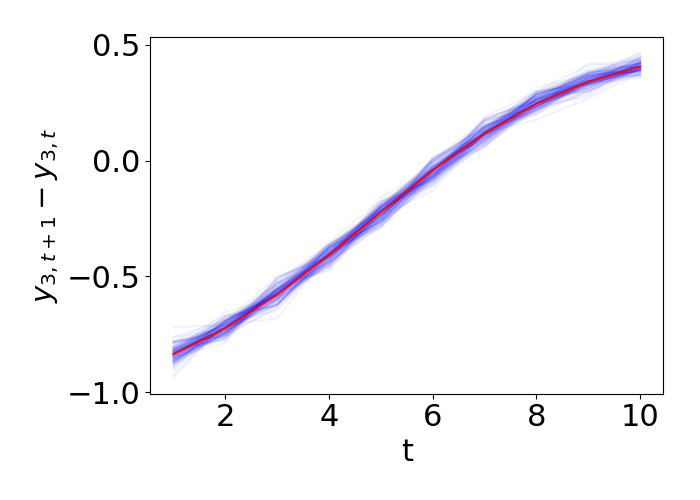}
    \label{fig:}
}
\subfigure[Matern32 kernel, $l=2$.]{
    \includegraphics[width=0.3\textwidth]{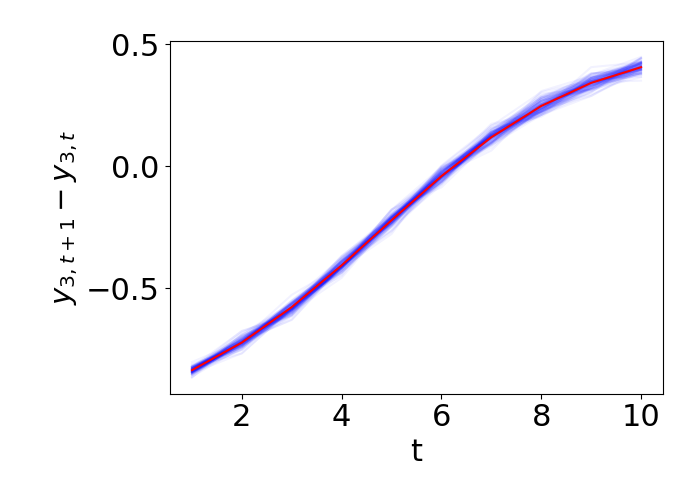}
    \label{fig:}
}
\subfigure[Matern32 kernel, $l=3$.]{
    \includegraphics[width=0.3\textwidth]{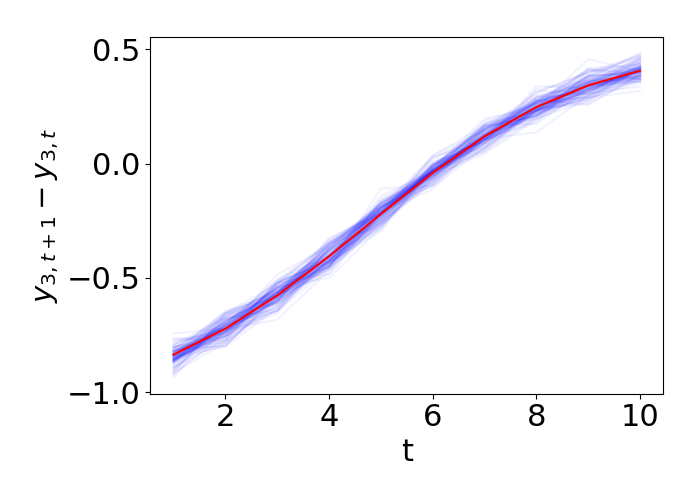}
    \label{fig:}
} \\
\subfigure[Matern12 kernel, $l=1$.]{
    \includegraphics[width=0.3\textwidth]{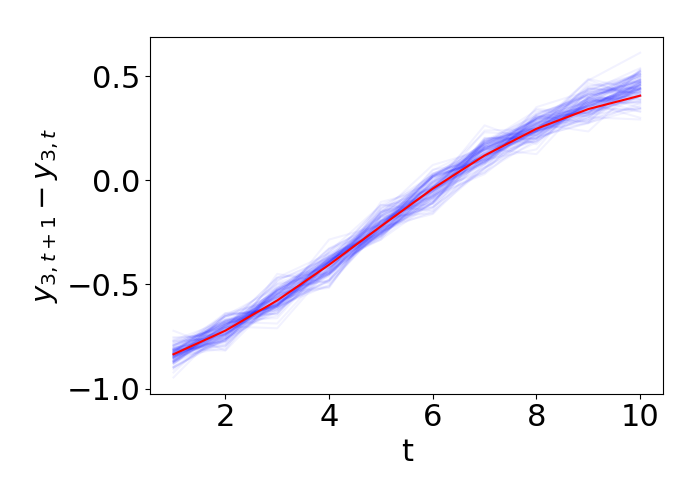}
    \label{fig:}
}
\subfigure[Matern12 kernel, $l=2$.]{
    \includegraphics[width=0.3\textwidth]{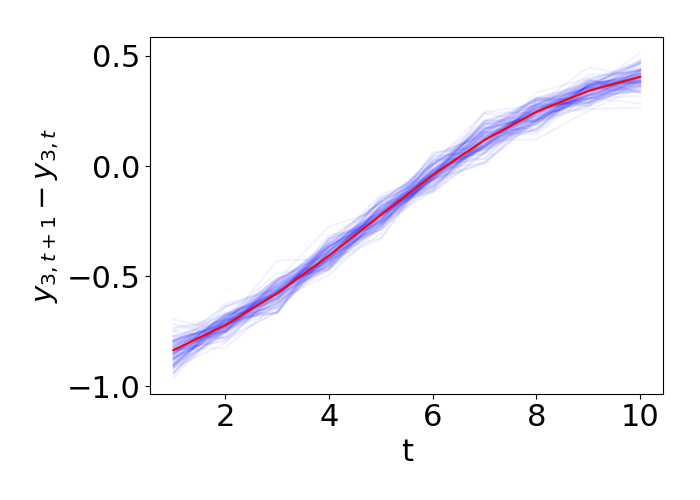}
    \label{fig:}
}
\subfigure[Matern12 kernel, $l=3$.]{
    \includegraphics[width=0.3\textwidth]{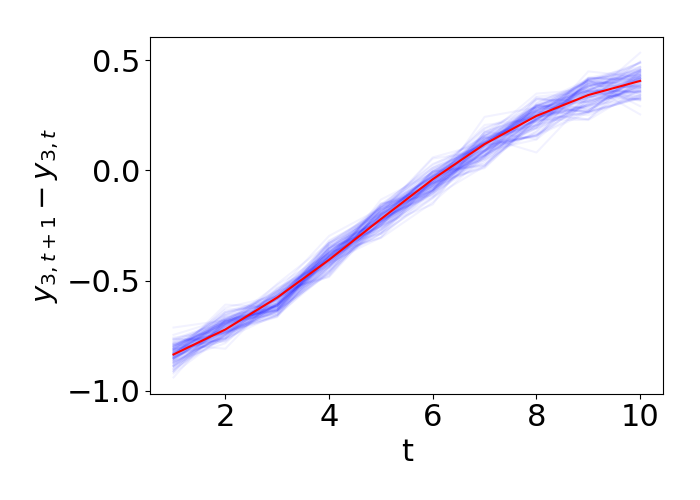}
    \label{fig:}
} \\
\caption{\textbf{Case b}. Analysis on the effect of kernel choice and depth when modelling the stationary dynamics in the modified cartpole example.
}
\label{fig:kernels_modified_cartpole_dynamics_no_crash}
\end{figure}

\paragraph{Case b} We train our DGP model on $50$ sequences of length $10$. Each sequence begins at a random initial point, under the same distribution used in our control experiments. Sequences of negative actions are then uniformly sampled in $[-3,0]$ and the resulting transitions are used to train our DGP model. This is then tested by sampling the trajectory under a sequence of fixed actions with $a=-1$ from the initial state mean, and comparing to the ground truth. See Figure~\ref{fig:kernels_modified_cartpole_dynamics_no_crash}.

\begin{figure}[ht!]
\centering
\subfigure[SEXP kernel, $l=1$.]{
    \includegraphics[width=0.3\textwidth]{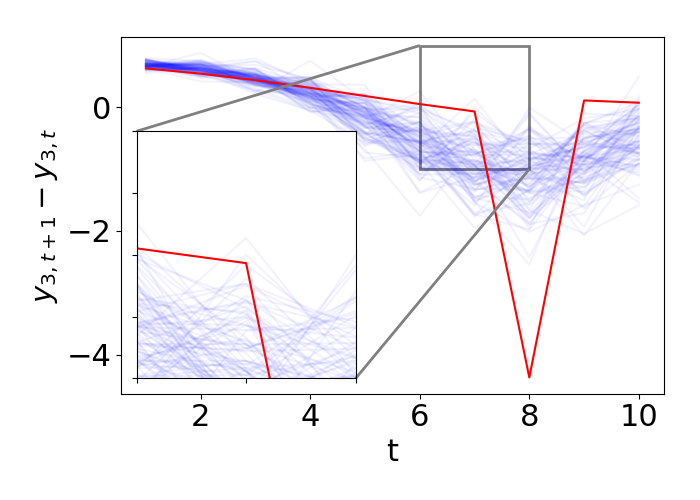}
    \label{fig:}
}
\subfigure[SEXP kernel, $l=2$.]{
    \includegraphics[width=0.3\textwidth]{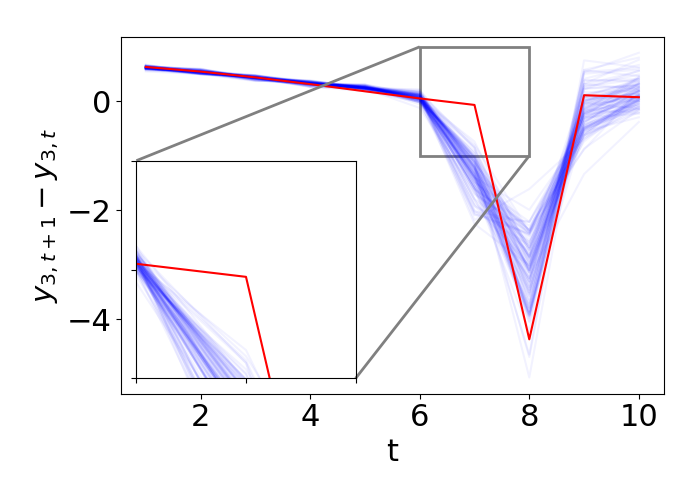}
    \label{fig:}
}
\subfigure[SEXP kernel, $l=3$.]{
    \includegraphics[width=0.3\textwidth]{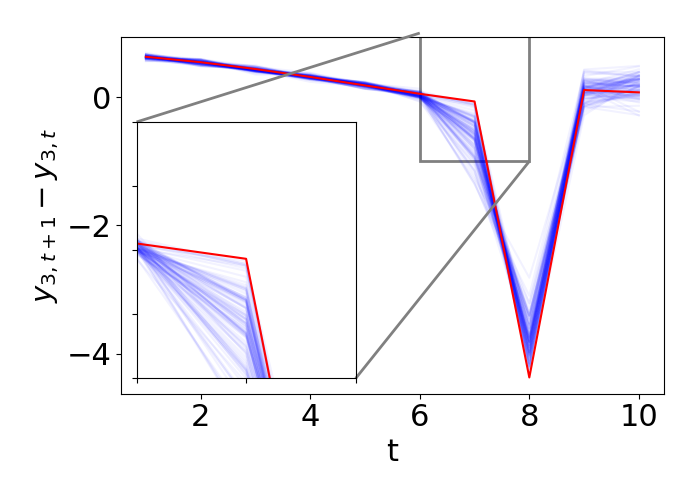}
    \label{fig:}
} \\
\subfigure[Matern52 kernel, $l=1$.]{
    \includegraphics[width=0.3\textwidth]{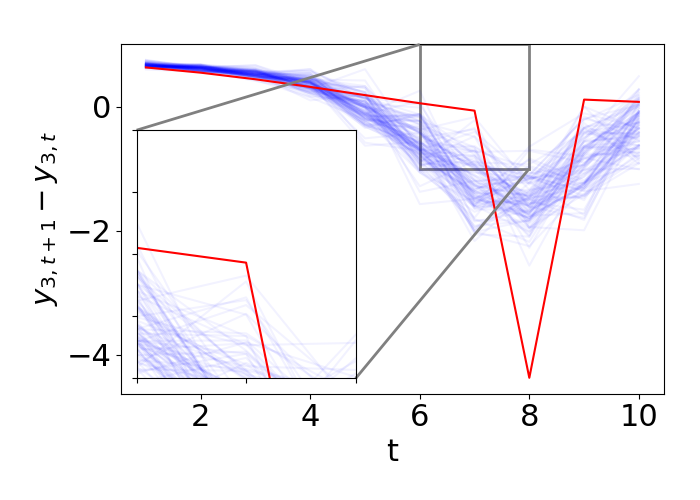}
    \label{fig:}
}
\subfigure[Matern52 kernel, $l=2$.]{
    \includegraphics[width=0.3\textwidth]{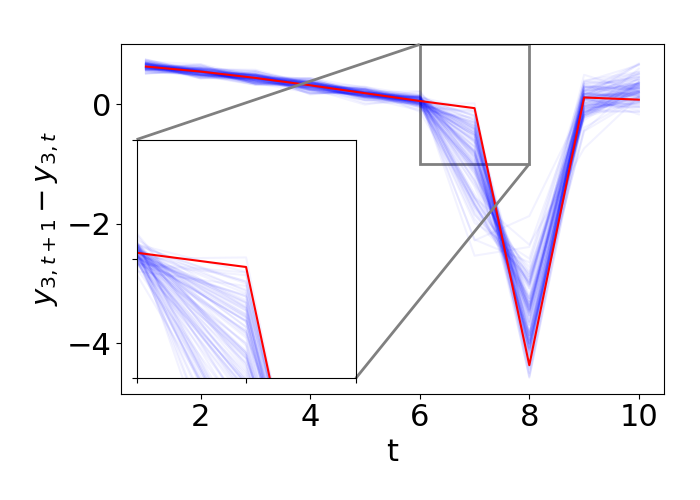}
    \label{fig:}
}
\subfigure[Matern52 kernel, $l=3$.]{
    \includegraphics[width=0.3\textwidth]{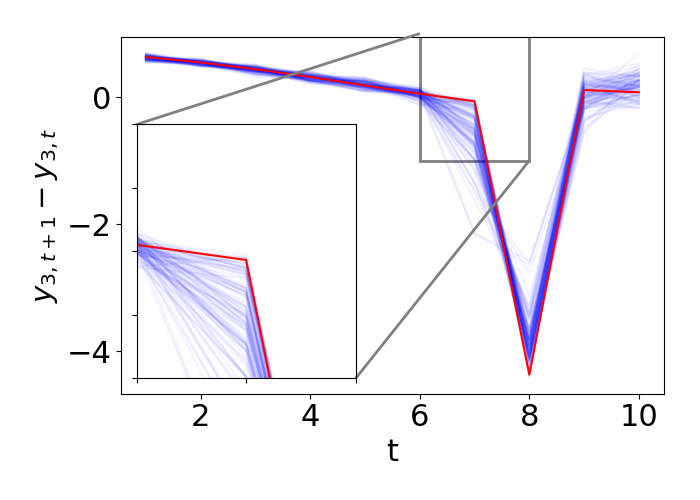}
    \label{fig:}
} \\
\subfigure[Matern32 kernel, $l=1$.]{
    \includegraphics[width=0.3\textwidth]{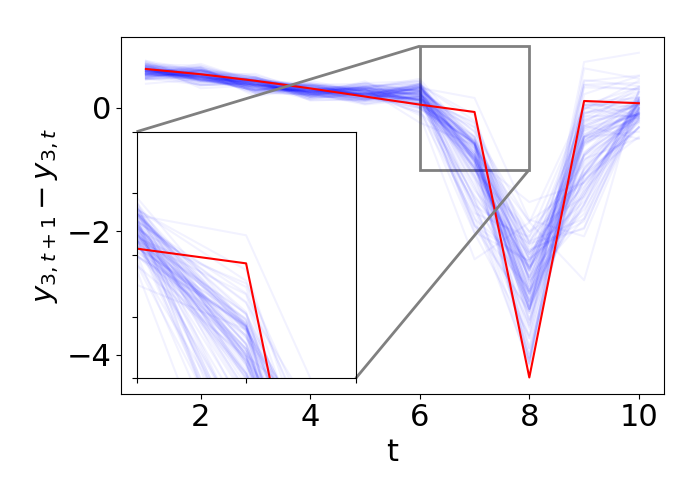}
    \label{fig:}
}
\subfigure[Matern32 kernel, $l=2$.]{
    \includegraphics[width=0.3\textwidth]{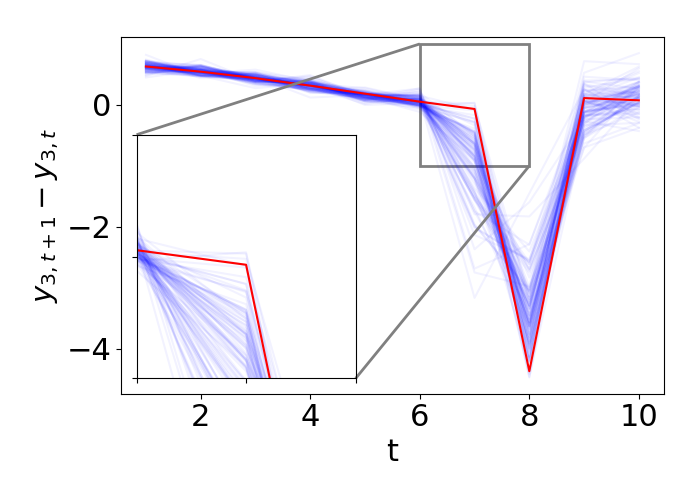}
    \label{fig:}
}
\subfigure[Matern32 kernel, $l=3$.]{
    \includegraphics[width=0.3\textwidth]{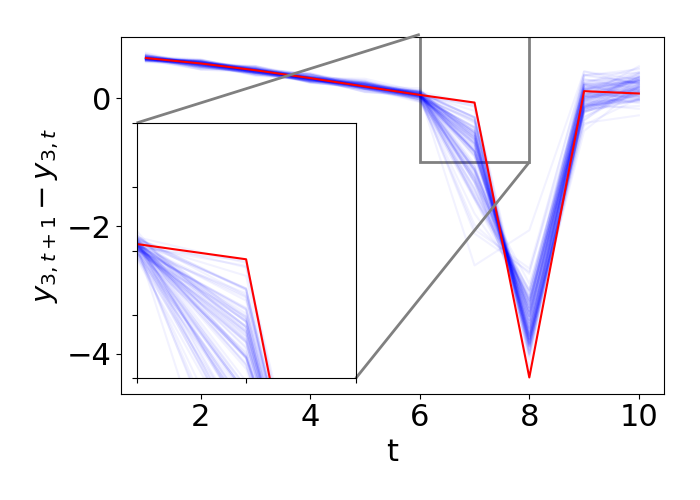}
    \label{fig:}
} \\
\subfigure[Matern12 kernel, $l=1$.]{
    \includegraphics[width=0.3\textwidth]{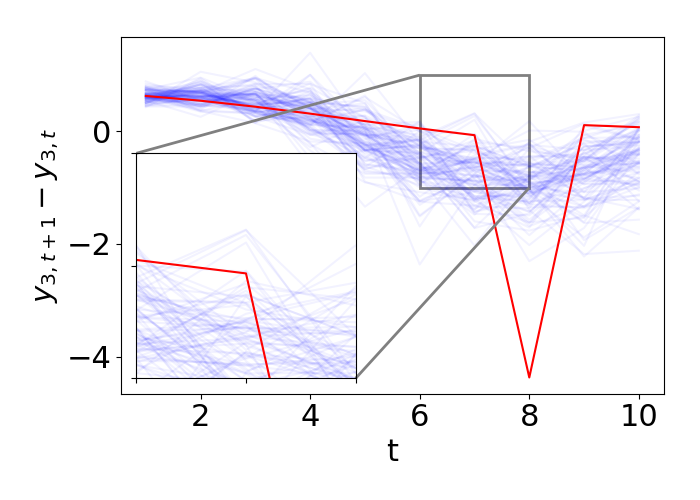}
    \label{fig:}
}
\subfigure[Matern12 kernel, $l=2$.]{
    \includegraphics[width=0.3\textwidth]{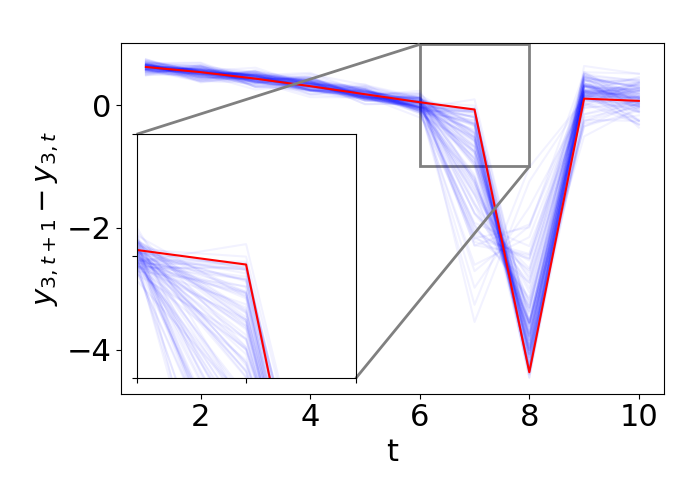}
    \label{fig:}
}
\subfigure[Matern12 kernel, $l=3$.]{
    \includegraphics[width=0.3\textwidth]{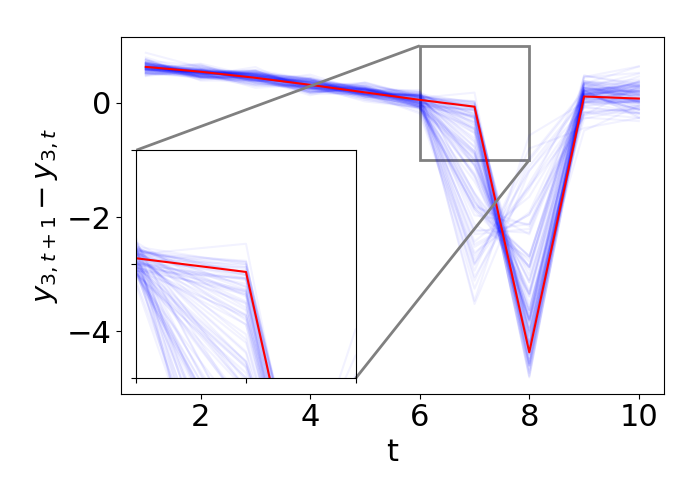}
    \label{fig:}
} \\
\caption{\textbf{Case c}. Analysis on the effect of kernel choice and depth when modelling both the stationary and non-stationary dynamics in the modified cartpole example.
}
\label{fig:kernels_modified_cartpole_dynamics_mixed}
\end{figure}

\paragraph{Case c} We train our DGP model on $50$ sequences of length $20$. Each sequence begins at a random initial point, under the same distribution used in our control experiments. Sequences of actions are then uniformly sampled in $[-3,3]$ and the resulting transitions are used to train our DGP model. This is then tested by sampling the trajectory under a sequence of fixed actions with $a=0.75$ from the initial state mean, and comparing to the ground truth. See Figure~\ref{fig:kernels_modified_cartpole_dynamics_mixed}.

\bibliographystyle{plainnat}
\bibliography{SM}

\vfill